\documentclass[lettersize,journal]{IEEEtran}
\usepackage[section]{placeins}
\usepackage{amsmath,amsfonts,amsthm,bm}
\usepackage{amssymb}
\usepackage{algorithmic}
\usepackage{algorithm}
\usepackage{array}
\usepackage{float}
\usepackage[caption=false,font=normalsize,labelfont=sf,textfont=sf]{subfig}
\usepackage{textcomp}
\usepackage{stfloats}
\usepackage{url}
\usepackage{color}
\usepackage{verbatim}
\usepackage{graphicx}

\newtheorem{prop}{Proposition}

\usepackage{multirow}
\definecolor{DarkGreen}{rgb}{0.2,0.5,0.2} 
\usepackage[%
pdftex,%
colorlinks=true,%
hyperindex,%
plainpages=false,%
pagebackref=true,%
bookmarksopen,%
bookmarksnumbered %
]{hyperref}
\graphicspath{{pic/}}
\definecolor{darkgreen}{rgb}{0,0.5,0}
\definecolor{purple}{rgb}{1,0,1}

\begin{document}
	
	\title{IDLS: Inverse Depth Line based Visual-Inertial SLAM }
	
	\author{Wanting Li, Shuo Wang, Yongcai Wang, Yu Shao, Xuewei Bai, Deying Li
		\thanks{W. Li, S. Wang, Y. Wang, Y. Shao, X. Bai, and D. Li are all with School of Information, Renmin University of China, Beijing, P.R.China, 100872}
		\thanks{Corresponding author: Yongcai Wang, ycw@ruc.edu.cn}
		\thanks{This work was supported in part by the National Natural Science Foundation of China Grant No. 61972404, 12071478, Public Computing Cloud, Renmin University of China}
		\thanks{Demo video is at \url{https://youtu.be/aGwxdrwk_QM}.}
	}
	
	
	
	\maketitle
	\thispagestyle{empty}
	\pagestyle{empty}
	\begin{abstract}
		For robust visual-inertial SLAM in perceptually-challenging indoor environments,  
  recent studies exploit line features to extract descriptive information about scene structure to deal with the degeneracy of point features. %
		But existing point-line-based SLAM methods mainly use Plücker matrix or orthogonal representation to represent a line, which needs to calculate at least four variables to determine a line. Given the numerous line features to  determine in each frame, the overly flexible line representation increases the computation burden and comprises the accuracy of the results. 
		In this paper, we propose inverse depth representation for a line, which models each extracted line feature using only two variables, i.e., the inverse depths of the two ending points. It exploits the fact that the projected line's pixel coordinates on the image plane are rather accurate, which partially restrict the line.  
		Using this compact line presentation, Inverse Depth Line SLAM (IDLS) is proposed to track the line features in SLAM in an accurate and efficient way.  A robust line triangulation method and a novel line re-projection error model are introduced.  And a two-step optimization method is proposed to firstly determine the lines and then to estimate the camera poses in each frame. 
		IDLS is extensively evaluated in multiple perceptually-challenging datasets.  The results show it is more accurate, robust, and needs lower computational overhead than the current state-of-the-art of point-line-based SLAM methods.
	\end{abstract}
	
	
	\section{Introduction}
	
	By combining visual information from cameras and inertial measurements from IMU, VI-SLAM (Visual-inertial Simultaneous Localization and Mapping) is a prevailing technique in robotics, Augmented Reality (AR), and Virtual Reality (VR) that enables a robot or a device to learn the map of an unknown environment and locates itself in that environment simultaneously\cite{kazerouni2022survey,gui2015review,ventura2014global,concha2021instant,ye2022coli}. 
	Traditionally, VI-SLAM methods heavily rely on point features for map representation and localization\cite{rublee2011orb}\cite{mur2015orb}\cite{mur2017orb}\cite{campos2021orb}. But in structurally similar and texture-less environments, such as corridors or parking lots,  the point feature-based VI-SLAM systems suffer from drifting problems and instability issues. 
	
	\begin{figure}[thbp]
		\centering
		\includegraphics[width=1\linewidth]{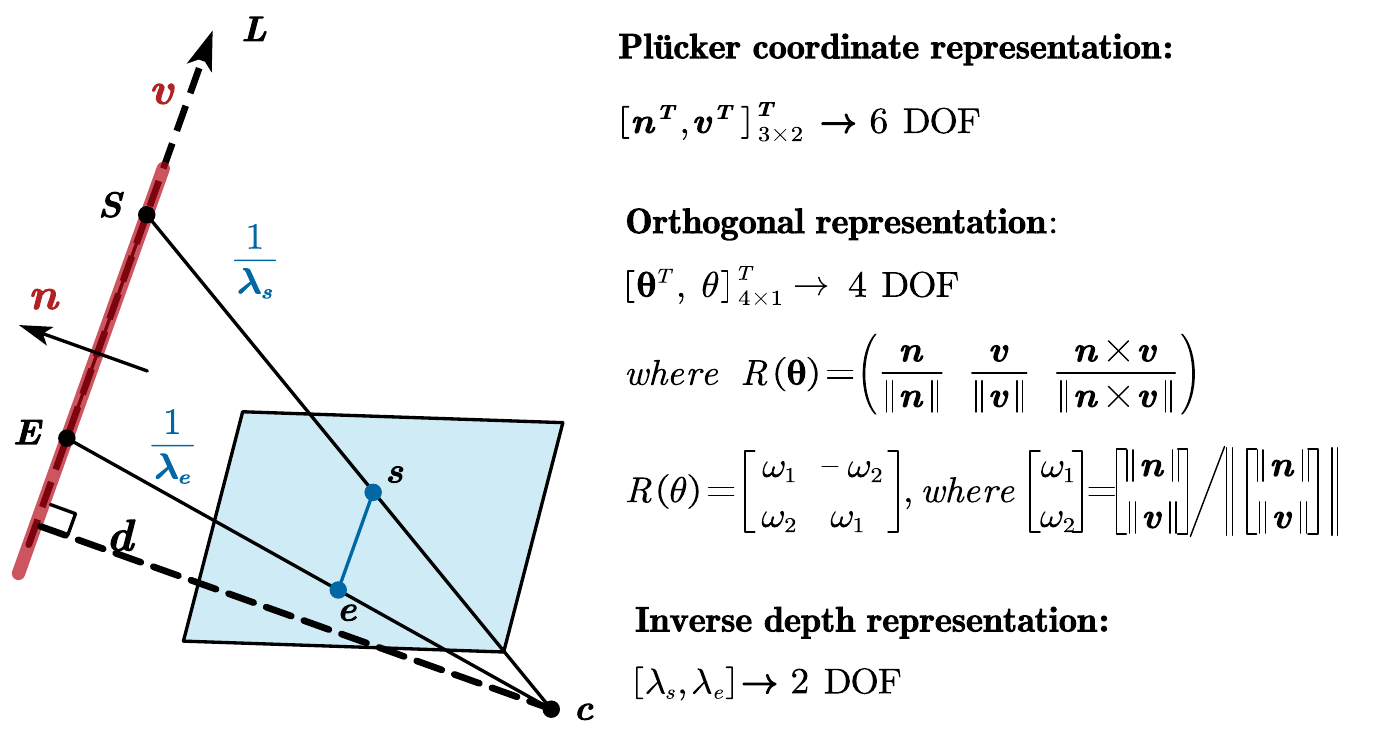}
		\caption{DOFs of different line representations. }
		\label{fig:line-model}
        \vspace{-0.2in}
	\end{figure}
	
	To enhance the performance of visual-inertial SLAM in perceptually challenging environments,  recent studies have explored line features to extract descriptive information about scene structures to enable the point-line-based VI-SLAM\cite{pumarola2017pl}\cite{he2018pl}\cite{yang2019visual}\cite{fu2020pl}\cite{zhou2021dplvo}\cite{lim2021avoiding}\cite{lim2022uv}. These methods improve SLAM system robustness and accuracy in challenging environments because there are rich line features in man-made indoor environments.

	However, existing point-line-based methods mainly use  Plücker coordinate representations\cite{Joswig2013} or orthogonal representations \cite{zhang2015building} to represent a line as shown in Fig.~\ref{fig:line-model}. 
	The Plücker coordinate representation needs six variables to represent a line, which is overly parameterized. The orthogonal representation presents a minimal representation of a line\cite{zhang2015building}, which still has four degrees of freedom.  
	Considering that there are numerous line features that need to be determined in each frame, the line representation with a slightly higher degree of freedom will significantly increase the computational burden and compromises the accuracy of the results.
	
	
    In fact, when a spatial 3D line is projected on the image plane, two degrees of freedom are fixed by the pixel coordinates\cite{civera2008inverse}, so the representation dimension can be further reduced. 
    Contemplating two points on the 2D line on the image plane, 
	only if the depths of their corresponding 3D points are calculated, can the corresponding 3D line be determined. 
	Therefore, we propose an inverse-depth representation for each 3D line, which represents each 3D line using two parameters, i.e., the inverse depths of two endpoints of a LSD (line segment detector) feature \cite{von2008lsd} detected in the first observed image.  Given the numerous line features in each frame, this dimension reduction line representation greatly reduces the variable dimension, therefore, increasing the computation efficiency and accuracy. 

	Based on the inverse-depth line representation, we introduce a novel triangulation method that avoids line direction parameter errors in line degenerate scenes. Furthermore, we propose a novel line reprojection error model that considers only the line's normal vector, making it more accurate in the camera pose optimization process.
	Also, because the two endpoints’ inverse depths are coupled with the camera pose, a two-step optimization method is proposed to enhance the optimization of the camera poses. %
	In each iteration, we first calculate the 3D lines using the fixed keyframe poses. Then we use the new line parameters to optimize the keyframe poses. Each optimization step is simple to solve, which improves the efficiency.
	
	By integrating the aforementioned line extraction and representation with Mono-camera-based SLAM method \cite{qin2018vins}, an \emph{Inverse Depth Line based SLAM method (IDLS)} is proposed. The key contributions include: 
	
	\begin{figure*}[!t]
		\centering
		\subfloat[]{\includegraphics[width=0.5\linewidth]{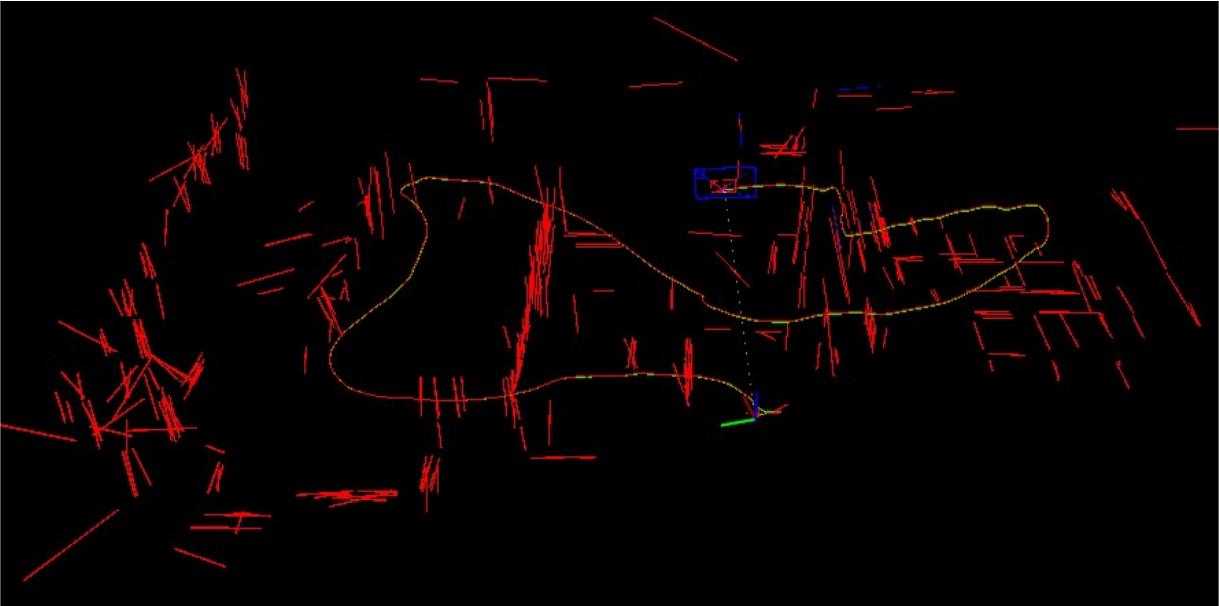}%
			\label{fig_first_case}}
		\subfloat[]{\includegraphics[width=0.39\linewidth]{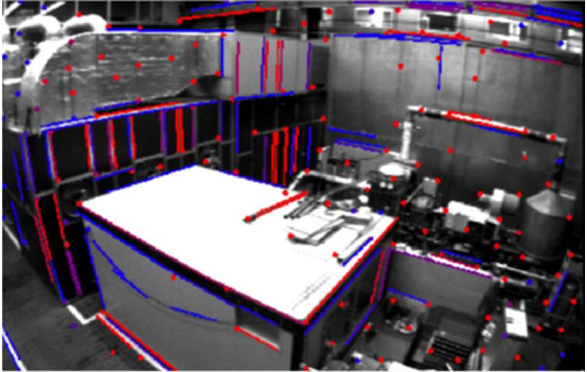}%
			\label{fig_second_case}}
		\caption{Figure shows the IDLS's trajectory and point-line map for the MH-04-Difficult sequence. The two images are screenshots of the ROS Rviz window, where the red lines represent tracked lines, the blue lines represent untracked lines, and the green line represents the motion trajectory. (a) SLAM trajectory and feature map window. (b) Feature tracked window.}
		\label{fig_result}
	\end{figure*}
	
	\begin{itemize}
		\item A two-parameter representation for a line based on the endpoints' inverse depth is presented.
		\item A robust triangulation method that avoids introducing the line direction parameter errors. 
		\item A novel line reprojection error model that considers only the line's normal vector in the camera pose optimization process. This approach allows a more accurate calculation of the line and the camera poses in real-time.
		\item A two-step optimization method that reduces the computational complexity, making each optimization step easier to solve.
		\item Above methods are integrated into IDLS, a monocular inertial SLAM algorithm that uses line and point features, which exhibits accuracy and efficiency in challenging environments. 
		\item Experiments in EuRoc\cite{burri2016euroc} and real perceptually challenging datasets demonstrate that IDLS outperforms state-of-the-art algorithms in a variety of challenging circumstances.
	\end{itemize}
	
\section{Related Work}
For the large body of research works in visual SLAM\cite{chen2020survey},  we focus on the most closely related point-line-based visual SLAM methods. 
Depending on the type of sensors used, monocular point-line-based SLAM (Simultaneous Localization and Mapping) can be roughly divided into two categories: VSLAM (Visual SLAM)\cite{gomez2016pl,pumarola2017pl,zhou2021dplvo,zhou2022edplvo} and VISLAM (Visual-Inertial SLAM)\cite{he2018pl,fu2020pl,lim2021avoiding,wei2021point,lee2021plf,liu2022plc,lim2022uv,xu2023eplf}. 
Each monocular point-line-based SLAM method mainly includes two phases: (1) the front-end feature extraction and matching; (2) the back-end pose optimization and loop closure detection.

\begin{table*}[]
	\caption{Point-Line SLAM\\(Blue color indicate that lines' DOF affect optimization, black color indicate that lines' DOF are only used to store features)}
	\resizebox{\textwidth}{!}{
	\begin{tabular}{lllllllll}
		\hline
		\multirow{3}{*}{Method} & \multirow{3}{*}{Sensors} & \multirow{3}{*}{Time} & \multicolumn{3}{l}{Front-end}                                                                                                                                                                 & \multicolumn{2}{l}{Back-end}                                 & \multirow{3}{*}{\begin{tabular}[c]{@{}l@{}}Loop \\ Closure\end{tabular}} \\
                        &                          &                       & \multirow{2}{*}{\begin{tabular}[c]{@{}l@{}}Line \\ Extraction\end{tabular}} & \multirow{2}{*}{\begin{tabular}[c]{@{}l@{}}Line \\ Tracking\end{tabular}} & \multirow{2}{*}{Line Triangulation} & \multicolumn{2}{l}{Line Reprojection Error}                  &                                                                          \\
                        &                          &                       &                                                                             &                                                                           &                                     & Model                                       & Representation &                                                                          \\\hline\hline
		PL-SVO\cite{gomez2016pl}                  & \multirow{4}{*}{VO}      & 2016&LSD                                                                         & Direct                                                                    & Endpoint photometric error            & 3D endpoint to 2D point                     &6DOF           &-                                                                         \\
		PL-SLAM\cite{pumarola2017pl}                 &                          &2017& LSD                                                                         & LBD                                                                       & 3D endpoint to 2D line                & 3D endpoint to 2D line                      &  \color{blue}6DOF           & \checkmark                                                                         \\
		DPLVO\cite{zhou2021dplvo}                  &                          &2021& LSD                                                                         & Direct                                                                    & Sampling Point Fitting               & collinear constraint                        & 2DOF           &-                                                                          \\
		EDPLVO\cite{zhou2022edplvo}                  &                          &2022& LSD                                                                         & Direct                                                                    & Sampling Point Fitting              & collinear constraint+line photometric error & 2DOF           &-                                                                          \\\hline
		PL-VIO\cite{he2018pl}                  & \multirow{8}{*}{VIO}     &2018& LSD                                                                         & LBD                                                                       & Plücker Matrix           & 3D endpoint to 2D line                      &  \color{blue}4DOF           &-                                                                         \\
		PL-VINS\cite{fu2020pl}               &                          &2020& LSD                                                                         & LBD                                                                       & Plücker Matrix                   & 3D endpoint to 2D line                      &  \color{blue}4DOF           &\checkmark                                                                          \\
		AD-SLAM\cite{lim2021avoiding}                   &                          &2021& LSD                                                                         & LBD                                                                       & Plücker Matrix                       & 3D endpoint to 2D line+parallel lines       &  \color{blue}4DOF           &\checkmark                                                                          \\
		PLF-VINS\cite{lee2021plf}               &                          &2021& LSD                                                                         & LBD                                                                       & Plücker Matrix                       & 3D endpoint to 2D line+parallel lines       &  \color{blue}4DOF           &\checkmark                                                                          \\
		PML-VIO\cite{wei2021point}                 &                          &2021& LSD                                                                         & Prediction                                                                & Plücker Matrix                       & 3D endpoint to 2D line                      &  \color{blue}4DOF           &\checkmark                                                                          \\
		PLC-VIO\cite{liu2022plc}                 &                          &2022& EDLine                                                                      & Geometric                                                                 & Plücker Matrix                      & 3D endpoint to 2D line                      &  \color{blue}4DOF           &\checkmark                                                                          \\
		UV-SLAM\cite{lim2022uv}                  &                          &2022& LSD                                                                         & LBD                                                                       & Plücker Matrix                       & 3D endpoint to 2D line+vanishing point      &  \color{blue}4DOF           &\checkmark                                                                          \\
		EPLF-VINS\cite{xu2023eplf}               &                          &2023& EDLine                                                                      & Opticalflow                                                               & Plücker Matrix                      & Angle between plane normal vectors          & \color{blue} 4DOF           &\checkmark 
		\\\hline                                                                        
	\end{tabular}
	}
	\label{tab:slam}
\end{table*}

\subsection{Front-end Approaches}

The front end generally contains the following steps to process the lines. (1) The feature lines are extracted; (2) tracked (direct-based method based on photometric errors and feature-based method based on descriptors);  and (3) triangulated to find the 3D poses.

\subsubsection{Line Extraction}
Many point-line-based SLAM (PL-SLAM) systems extract line features using LSD (Line Segment Detector)\cite{von2008lsd}. However, an issue of LSD is that the extracted line segments can be fragmented or broken. To address this problem, researchers have proposed different approaches to improve the quality of line features. PLC-VIO\cite{liu2022plc} proposes a new line segment merging method based on EDLines\cite{akinlar2011edlines} to extract long line segment features. EPLF-VINS\cite{xu2023eplf} improves the traditional line detection model based on EDLines to obtain high-quality line features by means of short line fusion and adaptive threshold extraction.

\subsubsection{Line Tracking}

Based on how the image frames are processed in the front end, line tracking methods can be classified into two categories: (1) direct methods\cite{gomez2016pl,zhou2021dplvo,zhou2022edplvo} and (2) feature-based (indirect) methods\cite{pumarola2017pl,he2018pl,fu2020pl,lim2021avoiding,wei2021point,lee2021plf,liu2022plc,lim2022uv,xu2023eplf}. 
Direct methods match features by minimizing the photometric error between two frames. PL-SVO\cite{gomez2016pl} minimizes the patched photometric error between frames for image alignment. 
DPLVO\cite{zhou2021dplvo} uses a method that uses angles and distances to the origin to represent lines and to seek to combine lines with similar parameters. 
EDPLVO\cite{zhou2022edplvo} adds a support domain to DPLVO for 
measuring the uncertainty of the line segments to optimize the search of collinear constraints.

Feature-based methods mainly use LBD descriptors\cite{zhang2013efficient} for line matching. To ensure reliability and stability, PML-VIO\cite{wei2021point} proposes a  "predictive-matching" line tracking method to increase the tracking length of line segments. PLC-VIO\cite{liu2022plc} improves feature line tracking based on geometric constraints between the line and the points on the line. EPLF-VINS\cite{xu2023eplf} proposes a fast method for line optical flow tracking based on the assumption of grey scale invariance and columnar constraints.

\subsubsection{Line Triangulation}
Line triangulation is to calculate the 3D pose of a line in space by observing the same line between multiple frames, which can also be divided into direct and feature-based two categories. 

The direct methods employ the Lucas-Kanade algorithm\cite{baker2004lucas} to minimize the photometric error between the current 2D line pose in the image and the projection of 3D line observations. 
PL-SVO\cite{gomez2016pl} uses two endpoints to represent a line. The photometric error of a line is represented by the sum of the photometric errors of the two endpoints when calculating the photometric error.
DPLVO\cite{zhou2021dplvo} and EDPLVO\cite{zhou2022edplvo} utilize point-to-line fitting approaches. These methods calculate the three-dimensional coordinates of points independently by minimizing the photometric error and then fit a line representation by calculating the line's Plücker coordinates.

For feature-based methods, PL-SLAM\cite{pumarola2017pl} uses observations in multiple frames to setup the residue function to optimize 3D line's Plücker coordinates. 
PL-VIO\cite{he2018pl} reduces the optimization complexity by using two frames of observation to obtain the Plücker coordinate representation of the line and solves the equation using the Plücker matrix method\cite{hartley2003multiple}.
But the two-frame-based triangulation is prone to errors. In this paper, we use multiple frame-based triangulation to minimize the reprojection error for obtaining better accuracy, while using low-dimension line representation to reduce the computation cost.

\subsection{Back-end Approaches}

The back-end module optimizes the poses of the features and the camera jointly by minimizing a square residue error function. Here, we focus on the increased line reprojection error contributed by the line features.

Direct methods use the photometric error of the line to set up the residual function. PL-SVO\cite{gomez2016pl} minimizes the reprojection error between the endpoints of the 3D line and the corresponding 2D line's endpoints in the image. 
DPLVO\cite{zhou2021dplvo} uses the distance from the 3D positions of the sampled points to the plane as the error function. 
EDPLVO\cite{zhou2022edplvo} uses the photometric error of the endpoints and the photometric error of the sampled points on the line segment and adds the collinear constraint.
Both DPLVO and EDPLVO represent the line using an angle and a distance in the 2D image plane, where the angle is between the line and the line connecting one endpoint and the origin; the distance is the line's distance to the origin. 
However, they calculate the line poses by fitting the sampling points. So the dimension-reduced line representation is not used in the line pose optimization process.

For feature-based methods, PL-SLAM\cite{pumarola2017pl} uses the sum of the distances from the projected points to the 2D line on the image plane as the residual error. 
The line representation is normalized Plücker coordinates (6DOF).
But the  Plücker coordinate representation is over-parameterized. Optimization of the line features can easily lead to a local optimum solution. From PL-VIO\cite{he2018pl} onwards, an orthogonal representation(4DOF) is used to reduce the dimension of the line representation. 
However, the line features are subject to degradation, so more constraints about line features have been introduced. 
AD-SLAM\cite{lim2021avoiding} identifies the degradation of lines using epipole positions and adds parallel line constraints. 
PLF-VINS\cite{lee2021plf} extracts coupled point-line features based on positional similarity and integrates the proposed parallel 3D line residuals into the optimization process. 
UV-SLAM\cite{lim2022uv} adds vanishing point identification and constraints. 
EPLF-VINS\cite{xu2023eplf} uses the angle between the plane normal vectors to measure the reprojection residuals of the line. 
Overall,  since PL-VIO, feature-based methods have mainly focused on modifying the line feature reprojection error function without altering the representation of line features to improve the optimization process. 



\subsection{Loop Closure}

The loop closure corrects long-term trajectory drift and is generally used in the feature-based method\cite{fu2020pl,lim2021avoiding, lee2021plf, lim2022uv, xu2023eplf}. The special is that the PL-VIO is a point and line odometer without a loop closure detection module, whereas the PL-VINS\cite{fu2020pl} complements the PL-VIO\cite{he2018pl}.

In overall, Table~\ref{tab:slam} summarizes the mostly related PL-SLAM methods. Existing PL-SLAM methods have been designed to improve SLAM accuracy, mainly by improving the extraction and tracking of lines at the front end and by introducing more constraints at the back end. 
The direct methods represent the lines using two DOFs but only for storage. The line poses are calculated by sampling point fitting.
The feature-based methods represent lines mainly using orthogonal representations which need four DOFs. 
We can conclude that if we intend to utilize multi-frame observations to conduct optimization-based accurate line triangulation, the degrees of freedom associated with line representation do have impacts on the accuracy and efficiency. Likewise, the back-end joint optimization is also impacted by the degrees of freedom of the line representation.
In this paper, we propose the inverse depth representation of lines, which reduces the variable dimensions of the problem in the optimization process to enhance the precision, robustness, and effectiveness of PL-SLAM.  \color{black}

	\section{Problem Model and Inverse Depth Line Geometry}
	We consider an autonomous agent, such as a UAV, a UGV, or a robot is moving in an GNSS denied indoor environment.  The agent is equipped with a monocular camera and an embedded IMU, which takes images in frequency $f_c$ and takes accelerator and gyroscope readings in frequency $f_i$ for estimating the agent's ego-motion and the surrounding environments simultaneously. 
	First, we establish the three coordinate frames for the world $\pi _w$, the IMU $\pi _b$, and the camera $\pi _c$. The $z$-axis of $\pi _w$ is parallel to the direction of gravity.
	
	For each incoming image that the camera has acquired, point and line features are detected and tracked. Point features are found using Shi-Tomasi\cite{shi1994good}, tracked using KLT\cite{baker2004lucas}, and inliers are found using RANSAC-based epipolar geometry constraint\cite{hartley2003multiple}. Regarding the line features, the LSD detector\cite{von2008lsd}  and LBD descriptor\cite{zhang2013efficient} are used to identify the line segments in the new frame.
	We match the line segments with LBD descriptors in two consecutive frames by KNN\cite{kaehler2016learning} to track them and remove outliers of line matches by geometric constraints. 
	We use $\{c_1, c_2, \cdots, \}$ to represent the camera poses at different time instances. 
	$F^i_{c_1}$ and $L^i_{c_1}$ are used to represent the $i$th point feature and the $i$th line feature captured by a camera at $c_1$ respectively. The superscript will be omitted
	if we refer to a specific point or a line. The transformation from the world frame to the camera frame at $c_1$ is denoted by $\mathbf{T}_{wc_1}=\left( \begin{matrix}
	\mathbf{R}_{wc_1}&		\mathbf{t}_{wc_1}\\
	\mathbf{0}^T&		1\\
	\end{matrix} \right) $. For the IMU, we pre-integrate new IMU measurements \cite{forster2016manifold} between two consecutive frames to update the newest body states. 
	We then introduce the different representation of the line features.

	\subsection{Line Feature Representation}
	Traditionally, a line in 3D space is  widely represented by six parameter Plücker coordinates\cite{Joswig2013}. In 2015, Zhang et al. \cite{zhang2015building} proposed four parameter orthonormal representation for 3D line for the purpose of efficient state optimization. The four parameter model of a 3-D line shows a superior performance in terms of accuracy and convergence in iterative line state optimization. 
%
%
	\subsubsection{Plücker Coordinate Representation}
	Consider a 3-D line is captured at two camera poses $c_1$ and $c_2$.
	We consider the line representation for ${L_{{c_1}}} \in {\pi _{{c_1}}}$ in the first camera's frame.  The Plücker coordinates describe the line with ${L_{{c_1}}} = (\mathbf n_{_{{c_1}}}^\top, \mathbf d_{_{{c_1}}}^\top)^\top \in {\mathbb{R}^6}$, where $\mathbf n_{{c_1}} \in {\mathbb{R}^3}$ denotes the plane normal vector determined by $L_{{c_1}}$ and the origin of ${\pi _{{c_1}}}$. $\mathbf d_{{c_1}} \in {\mathbb{R}^3}$ denotes the direction vector determined by the two endpoints of ${L_{{c_1}}}$. 
	The Plücker coordinates are over-parameterized with a constraint $\mathbf n_{_{{c_1}}}^\top \mathbf d_{_{{c_1}}}=0$, so their dimension can be further reduced. 
	
	\subsubsection{Orthonormal Representation}
	Zhang et al. demonstrated that the Plücker coordinates can be described using a minimal four-parameter orthonormal representation\cite{zhang2015building}. According to the orthonormal representation,  $(\mathrm U, \mathrm W) \in SO(3) \times SO(2)$ can be calculated from ${L_{{c_1}}} = (\mathbf n_{_{{c_1}}}^\top, \mathbf d_{_{{c_1}}}^\top)^\top$ as: 
	\begin{equation}
	\mathrm{U}=\left[ \frac{\mathbf{n}_{c_1}}{\left\| \mathbf{n}_{c_1} \right\|}\,\, \frac{\mathbf{d}_{c_1}}{\left\| \mathbf{d}_{c_1} \right\|}\,\, \frac{\mathbf{n}_{c_1}\times \mathbf{d}_{c_1}}{\left\| \mathbf{n}_{c_1}\times \mathbf{d}_{c_1} \right\|} \right] 
	\end{equation}
	\begin{equation}
	\mathrm{W}=\left[ \begin{matrix}
	\omega _1&	\!\!	-\omega _2\\
	\omega _2&	\!\!	\omega _1\\
	\end{matrix} \right] , \! where \! \,\,\left[\!\! \begin{array}{c}
	\omega _1\\
	\omega _2\\
	\end{array}\!\! \right] \!\!=\!\!{{\left[\!\! \begin{array}{c}
			\left\| \mathbf{n}_{c_1} \right\|\\
			\left\| \mathbf{d}_{c_1} \right\|\\
			\end{array} \!\!\right]}\Bigg/{\left\| \left[\!\! \begin{array}{c}
			\left\| \mathbf{n}_{c_1} \right\|\\
			\left\| \mathbf{d}_{c_1} \right\|\\
			\end{array} \!\!\right] \right\|}}
	\end{equation}
	
	In practice, $\mathrm U, \mathrm W$ can be computed using the QR decomposition of $\left[ \begin{matrix}
	\mathbf{n}_{c_1}&		\mathbf{d}_{c_1}\\
	\end{matrix} \right] $: 
	\begin{equation}
	\left[ \begin{matrix}
	\mathbf{n}_{c_1}&		\mathbf{d}_{c_1}\\
	\end{matrix} \right] = \mathrm U \left[ {\begin{array}{*{20}{l}}
		{{\omega _1}}&0 \\ 
		0&{{\omega _2}} \\ 
		0&0 
		\end{array}} \right], and~  \mathrm W = \left[ {\begin{array}{*{20}{c}}
		{{\omega _1}}&{{\omega _2}} \\ 
		{ - {\omega _2}}&{{\omega _1}} 
		\end{array}} \right]
	\end{equation}
	where $\mathrm U$ and $\mathrm W$ denote a three-dimensional and a two-dimensional rotation matrix $R(\bm{\theta})$ and $R(\theta)$ respectively. So the line is essentially represented by four parameters $\mathbf p^\top =(\bm{\theta}^\top, \theta)$. Converting from the orthonormal representation to Plücker coordinates is easy, which is:
	\begin{equation}
	\left[ \begin{matrix}
	\mathbf{n}_{c_1}&		\mathbf{d}_{c_1}\\
	\end{matrix} \right] =\left[ \omega _1\mathbf{u}_{1}^{\top},\omega _2\mathbf{u}_{2}^{\top} \right] 
	\end{equation}
	where $\mathbf{u}_i$ is the $i$th column of $\mathrm U$.
	
	Although the orthonormal representation of line reduces dimension and can speed up convergence, both it and the Plücker coordinates 
	haven't considered the context provided by the projected line in the captured image.

	\subsubsection{Inverse Depth Representation}
\begin{figure}[!t]
		\centering
		\includegraphics[width=0.8\linewidth]{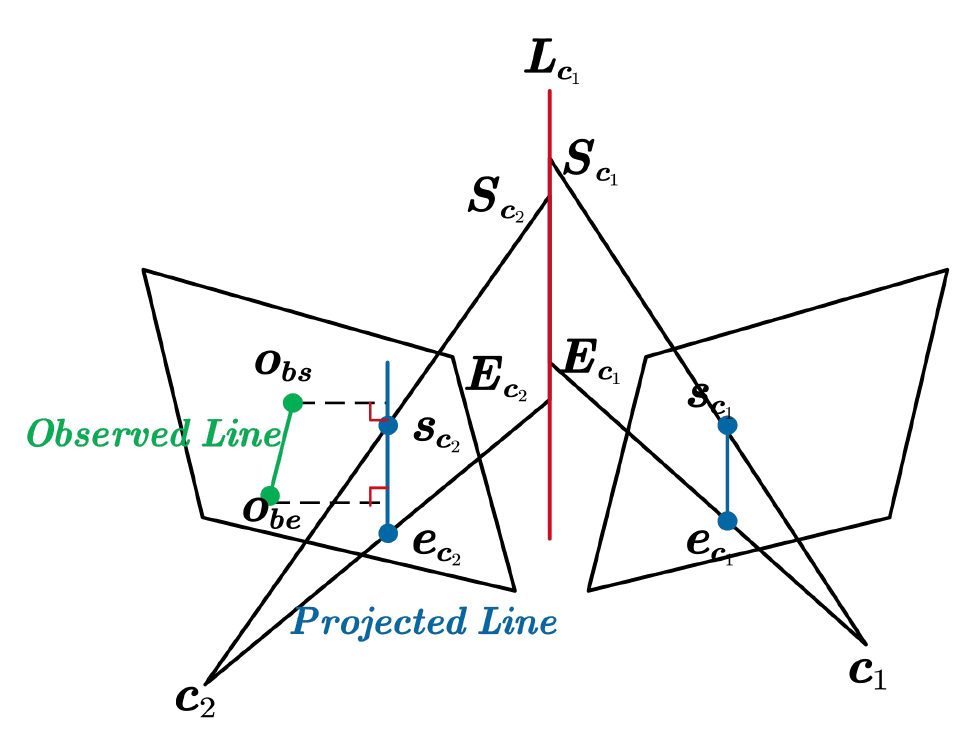}
		\caption{The line reprojection residual of an observed line in a sliding window is modelled in terms of the point-to-line distance.}
		\label{fig:line-residual}
        \vspace{-0.3in}
	\end{figure}

	Because two points determine a straight line, we exploit the inverse depth representation of two end  points to further reduce the parameter dimension of a 3-D line. Considering a 3-D line $L_{c_{1}}$   captured by the camera $c_1$. 
	Suppose two points on the line under the camera  $c_1$'s coordinate system are $S_{c_1}=(x_s^{{c_1}},y_s^{{c_1}},z_s^{{c_1}})$ and $E_{c_1}=(x_e^{{c_1}},y_e^{{c_1}},z_e^{{c_1}})$. They are represented in the normalized camera coordinate system as $s_{c_1}=(u_s^{{c_1}},v_s^{{c_1}},1)$ and $e_{c_1}=(u_e^{{c_1}},v_e^{{c_1}},1)$ as shown in Fig.\ref{fig:line-residual}. In the same way as the space point, the following equation can be obtained.
	
	\begin{equation}
	\begin{gathered}
	S_{c_1}=(x_s^{{c_1}},y_s^{{c_1}},z_s^{{c_1}}) = \frac{1}{{{\lambda _s}}}(u_s^{{c_1}},v_s^{{c_1}},1)=\frac{1}{{{\lambda _s}}}s_{c_1} \hfill \\
	E_{c_1}=(x_e^{{c_1}},y_e^{{c_1}},z_e^{{c_1}}) = \frac{1}{{{\lambda _e}}}(u_e^{{c_1}},v_e^{{c_1}},1)=\frac{1}{{{\lambda _s}}}e_{c_1} \hfill \\ 
	\end{gathered} 
	\end{equation}

	where $\lambda _s$ and $\lambda _e$ are the inverse depth of the two endpoints. According to the geometric principle, the normal vector $\mathbf n_{{c_1}}$ is the cross product of the vectors formed by the origin and the two points, and the direction vector $\mathbf d_{{c_1}}$ is the vector obtained by subtracting the two points. Because the endpoints' projected pixels on the image plane, i.e., $(u_s^{{c_1}},v_s^{{c_1}})$, $(u_e^{{c_1}},v_e^{{c_1}})$ are known, given two inverse depth parameters $l=(\lambda _s,\lambda _e)$, we can recover the  Plücker coordinates by:
	\begin{equation}
	\begin{gathered}
	\mathbf n_{{c_1}} = \frac{1}{{{\lambda _s}}}(u_s^{{c_1}},v_s^{{c_1}},1) \times \frac{1}{{{\lambda _e}}}(u_e^{{c_1}},v_e^{{c_1}},1)\hfill \\
	\mathbf d_{{c_1}} = \frac{1}{{{\lambda _e}}}(u_e^{{c_1}},v_e^{{c_1}},1) - \frac{1}{{{\lambda _s}}}(u_s^{{c_1}},v_s^{{c_1}},1)\hfill \\ 
	\end{gathered} 
	\label{eqn:line1}
	\end{equation}

	In this way we have reduced the line representation from 6 DOF (degrees of freedom) to 2 DOF, i.e., $\lambda _s$ and $\lambda _e$. 
	Comparing with the orthonormal representation which uses 4 degrees of freedom to optimize, the inverse depth representation utilizes the knowledge that the endpoints' projected coordinate on the camera plane are known and are rather accurate. The uncertainty of endpoints depends on the resolution of the image. So that the line is restricted in the plane formed by the projected line and the camera optical center. 
	The two endpoints depth value $d_s, d_e$, and so that the reverse depth parameters $\lambda _s=\frac{1}{d_s}$, $\lambda _e=\frac{1}{d_e}$ are the key parameters to determine the 3-D line in this scenario. 
	Because there are many lines detected in each frame, by reducing each line's degree of freedom, the problem's state dimension is greatly reduced. This enables better overall accuracy and efficiency in PL-SLAM.  
	
	\vspace{-0.2in}
	\subsection{Line's Reprojection Residual Error Model}

	Based on the inverse depth line representation model, we firstly introduce the line's reprojection residual error model in the back-end graph optimization process.  
	PL-SLAM sets up the square residual error minimization objective and solves the graph optimization problem \cite{leutenegger2015keyframe} to estimate the camera's ego-motion and the poses of points and lines. The objective function is mainly composed by: (1) the point-based reprojection residual error; (2)  the IMU's pre-integration residual error;  and (3) the line-based reprojection residual error. Since the point-based and the IMU pre-integration residual errors have been thoroughly introduced in \cite{qin2018vins}, in this subsection, we only introduce the lines' reprojection residual error model using the inverse depth representation.

	We model the line re-projection residuals based on the distance from the point to line as shown in Fig.\ref{fig:line-residual}. At first, we define line geometry transformation. 
	Let ${\mathbf R_{w{c_1}}},{\mathbf R_{w{c_2}}} \in SO(3)$ and ${\mathbf t_{w{c_1}}},{\mathbf t_{w{c_2}}} \in \mathbb{R}^3$ be the rotation matrices and translation vectors from $\pi_w$ to $\pi_{c_1}$, and from $\pi_{w}$ to $\pi_{c_2}$ respectively.  With these matrices, 
	by knowing the line $L_{c_1}$ captured by $c_1$, represented by (\ref{eqn:line1}),  
	we can transform $L_{c_1}$ in $\pi _{c_1}$ to $L_{c_2}$ in $\pi _{c_2}$ by the line transformation equation \cite{Joswig2013}:
	\begin{equation}
	\begin{gathered}
	L_w\!=\!\left[ \!\begin{array}{c}
	\mathbf{n}_w\\
	\mathbf{d}_w\\
	\end{array}\! \right] \!=\!\mathbf{T}_{wc_1}L_{c_1}\!=\!\left[ \!\begin{matrix}
	\mathbf{R}_{wc_1}\!\!&	\!\!\left[ \mathbf{t}_{wc_1} \right] _{\times}\mathbf{R}_{wc_1}\\
	0\!\!&\!\!		\mathbf{R}_{wc_1}\\
	\end{matrix}\! \right] \left[ \!\begin{array}{c}
	\mathbf{n}_{c_1}\\
	\mathbf{d}_{c_1}\\
	\end{array}\! \right]  \hfill \\
	L_{c_2}=\left[ \begin{array}{c}
	\mathbf{n}_{c_2}\\
	\mathbf{d}_{c_2}\\
	\end{array}\! \right] =\mathbf{T}_{c_2w}L_w={\mathbf{T}_{wc_2}}^{-1}L_w \hfill \\
	=\left[ \begin{matrix}
	\mathbf{R}_{wc_2}^{\top}\!\!& \!\!-{\mathbf{R}_{wc_2}}^{\top}\left[ \mathbf{t}_{wc_2} \right] _{\times}\\
	0\!\!&\!\! {\mathbf{R}_{wc_2}}^{\top}\\
	\end{matrix} \right] \left[ \begin{matrix}
	\mathbf{R}_{wc_1}\!\!&\!\!\left[ \mathbf{t}_{wc_1} \right]_{\times} \mathbf{R}_{wc_1}\\
	0\!\!&\!\!\mathbf{R}_{wc_1}\\
	\end{matrix} \right] \left[\!\!\begin{array}{c}
	\mathbf{n}_{c_1}\\
	\mathbf{d}_{c_1}\\
	\end{array} \!\!\right]   \hfill \\ 
	\end{gathered}
	\label{eqn:plucker-transform} 
	\end{equation}
	
	Next, the estimated projection line $\mathbf l_{c_2}$ on $c_2$'s image frame is obtained by transforming $L_{c_2}$ to the image plane\cite{zhang2015building}:
	\begin{equation}
	\mathbf l_{c_2} = \left[l_1,l_2,l_3\right]^\top = {\mathbf K_L}{\mathbf n_{{c_2}}}
	\label{eqn:l}
	\end{equation}
	where ${\mathbf K_L} = \left[ {\begin{array}{*{20}{c}}
		{{f_y}}&0&0 \\ 
		0&{{f_x}}&0 \\ 
		{ - {f_y}{c_x}}&{ - {f_x}{c_y}}&{{f_x}{f_y}} 
		\end{array}} \right]
	$ denotes the line projection matrix and $\mathbf n_{c_2}$ can be extracted from (\ref{eqn:plucker-transform}). Note that $L_{c_1}=[\mathbf n_{c_1}^\top, \mathbf d_{c_1}^\top]^\top$ is determined by the two inverse depth parameters in (\ref{eqn:line1}).

	Finally, assume that $L^i_{c_1}$ corresponds to the $i$-th matched line feature which is simultaneously observed by $c_1$ and $c_2$, denoted by $\mathbf {l}_{c_1}^i$ on $c_1$ and $\mathbf l_{c_2}^i = \left[l_1^{c_2}, l_2^{c_2}, l_3^{c_2}\right]^\top$ on $c_2$ as given in (\ref{eqn:l})  (a superscript is added to indicate the camera pose index). 
	Suppose the measured endpoints of this line on $c_2$ image plane is $s_{c_2}$ and $e_{c_2}$ respectively. Then this line's re-projection error can be calculated by the point to line distance.

	\begin{equation}
	r_l(z_{\mathbf {l}^i_{c_1}}^{c_1},\mathcal{X} )=\left[ \begin{array}{c}
	d(s_{c_2},\mathbf {l}^i_{c_2})\\
	d(e_{c_2},\mathbf {l}^i_{c_2})\\
	\end{array} \right] =\left[ \begin{array}{c}
	\frac{{s_{c_2}}^{\top}\mathbf {l}^i_{c_2}}{\sqrt{{l_{1}^{c_2}}^{2}+{l_{2}^{c_2}}^{2}}}\\
	\frac{{e_{c_2}}^{\top}\mathbf {l}^i_{c_2}}{\sqrt{{l_{1}^{c_2}}^{2}+{l_{2}^{c_2}}^{2}}}\\
	\end{array} \right] 
	\label{eqn:lineresidue}
	\end{equation}
	where $d({s_{{c_2}}},\mathbf {l}^i_{c_2})$ and $d({e_{{c_2}}},\mathbf {l}^i_{c_2})$ denote the point-to-line distance function. 
	In calculating the residual error, an important property should be noted. 
	Although we use the inverse depth of the two endpoints to represent the line feature, we use the distances from the observed endpoints  to the projected line, i.e., point-to-line distance to represent the reprojection error.  This doesn't require the same line feature captured in the two frame have the common end points. 
	
	\begin{prop}
		In calculating the frame-to-frame line re-projection residual error using the matched line segments $L_{c_1}$ and $L_{c_2}$ captured by $c_1$ and $c_2$ respectively, these two captured line segments don't need to have common 3D ending vertices. 
	\end{prop}

  As can be seen from Fig.\ref{fig:line-residual}, even if the observed endpoints $S_{c_1}, E_{c_1}$'s world coordinates  are different from the world coordinates of $S_{c_2}, E_{c_2}$, the line re-projection error in (\ref{eqn:lineresidue}) can be correctly calculated,  because the residual error is calculated by the point-to-line distance instead of endpoint-to-endpoint distance.  So $L_{c_1}$ and $L_{c_2}$ don't need to observe the common 3D endpoints of the line. 

	This property gives high flexibility in constructing the line re-projection residual errors using LSD-based line features, since the LSD detector provide two endpoints of the line in the image frame\cite{von2008lsd}. No matter whether the endpoints are from common points, only if they are from the same line, the residual error can be successfully constructed.  
	

    Considering the state variables in an optimized sliding window defined in (\ref{eqn:state}). Suppose there are $n$ keyframes in the sliding window, containing $n_p$ point features, and $n_l$ line features in total. Then the number of variables to estimate in the sliding window is:
    $v_n=12 \times n + n_p +2\times n_l$ using inverse depth line model and is $v_n=12 \times n + n_p +4\times n_l $ using the orthogonal representation.  For a typical case when $n=10, n_p=100, n_l=100$, the inverse depth representation has a $\frac{420}{620} \approx 2/3$ reduction on the number of variables than that of the orthogonal representation. The reduction is significant considering the $O(v_n^2)$ complexity in each iteration of optimization. In Section~\ref{sec:two-step}, we design a two-step optimization method to further speed up and to improve the accuracy. 
    

	Knowing the residual error function,  we need to further calculate the corresponding Jacobian matrix $J$ for updating the line parameter $\lambda_e$ and $\lambda_s$. The derivation of the Jacobian matrix is given in Appendix. 
	
	\subsection{Inverse Depth Line State Initialization}
	After setting the graph optimization model, the state variables needs to be initialized before conducting the iterative optimization. The camera poses can be initialized by IMU integration and the feature point locations can be initialized by triangulation method\cite{qin2018vins}.  
	In this subsection, we mainly introduce how to initialize the two endpoints inverse depths $\lambda _s,\lambda _e$ for each line segment. We propose an optimization-based method that triangulates the line parameters using multiple frames' information. It avoids the effect of one frame's error. This multi-frame optimization method can also avoid the influence of line degradation scenes, such as the transformations containing only rotations. 
    
	\begin{figure}[htbp]
		\centering
		\includegraphics[width=0.7\linewidth]{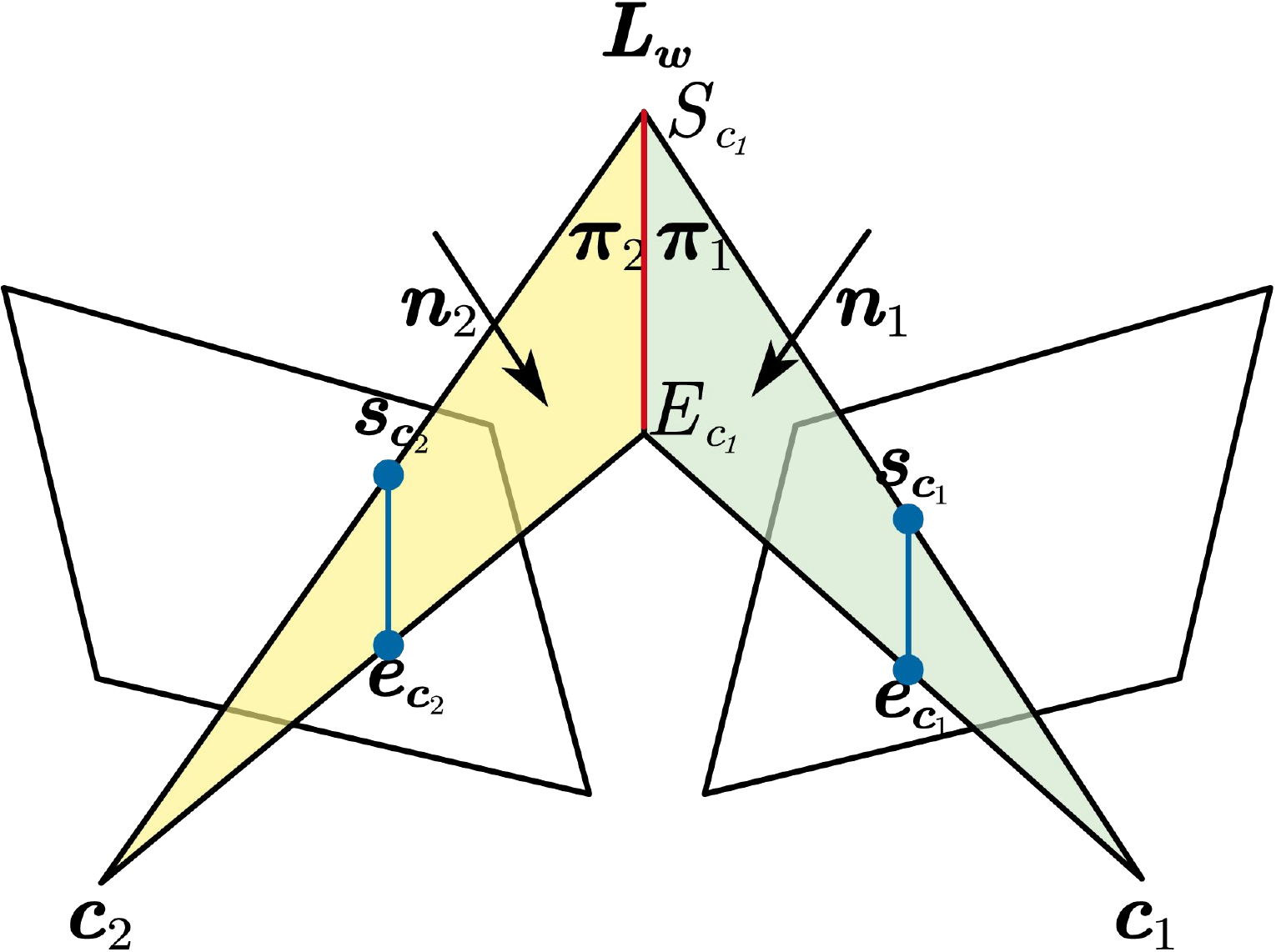}
		\caption{Initialization of a newly observed line. The straight line $L_{c_1}$ can be regarded as the intersection of two planes $\pi_1$ and $\pi _2$ to obtain, that is, this line is in both planes and the distance to both planes is 0.}
		\label{fig:line-representation}
	\end{figure}
	
	Consider the instance showing in Fig.\ref{fig:line-representation} 
	when a line is captured by two camera frames.   
	let's assume a line $L_{w}$ is observed by two camera poses $c_1$ and $c_2$ respectively. Then two planes are determined by the two camera centers and $L_{w}$, which are denoted $\pi _1 = (c_1, L_{w})$ and $\pi _2 = (c_2, L_{w})$. The line $L_{w}$ is the intersection of the two planes, so its distance to both planes is 0. 
	The normal vector of the plane $\pi_2$ in $c_2$'s coordinate system can be represented by $\mathbf n_2=s_{c_2}\times e_{c_2}$, which is the cross product of the two vectors on the plane. Then the normal vector of the plane $\pi_2$ in $c_1$'s coordinate system
	can be represented by $\mathbf n_2^{c_1}= \mathbf R_{{c_1}{c_2}}s_{c_2}\times e_{c_2}$, where $\mathbf R_{{c_1}{c_2}}$ is the rotation matrix from $\pi _{c_1}$ to $\pi _{c_2}$. Then plane $\pi_2$ can be expressed in point-normal form as $\pi _2=\{ {{\mathbf n_2}^{c_1}}, - {{\mathbf n_2}^{c_1}}\cdot {c_2}\}$. We can then convert $\pi _2$'s representation to its general representation \cite{plane} which has four parameters $(\pi _{A}^{c_2},\pi _{B}^{c_2},\pi _{C}^{c_2}, \pi _{D}^{c_2})$: where
	\begin{equation}
	\pi _{A}^{c_2},\pi _{B}^{c_2},\pi _{C}^{c_2}=\mathbf{n}_{2}^{c_1}, \pi _{D}^{c_2}=-\mathbf{n}_{2}^{c_1}\cdot c_2
	\label{eqn:normalpi2}
	\end{equation}
	
	The observation of $L_w$ is $L_{c_1}$ in $c_1$'s coordinate frame. We then conduct following analysis in $c_1$'s coordinate frame. 
	The two endpoints of $L_{c_1}$, i.e., $S_{c_1}$ and $E_{c_1}$ are on $L_{c_1}$, which is on plane $\pi_2$. So the distance from each endpoint to the plane $\pi _2$ is $0$. So we get the two endpoints' inverse depth $\lambda _s,\lambda _e$ from the following equation.
	
	\begin{equation}
	\begin{array}{c}
	d\left( S_{c_1},\pi _2 \right) =\frac{|(\pi _{A}^{c_2},\pi _{B}^{c_2},\pi _{C}^{c_2})\cdot S_{c_1}+\pi _{D}^{c_2}|}{\sqrt{{\pi _{A}^{c_2}}^2+{\pi _{B}^{c_2}}^2+{\pi _{C}^{c_2}}^2}}=0\\
	d\left( E_{c_1},\pi _2 \right) =\frac{|(\pi _{A}^{c_2},\pi _{B}^{c_2},\pi _{C}^{c_2})\cdot E_{c_1}+\pi _{D}^{c_2}|}{\sqrt{{\pi _{A}^{c_2}}^2+{\pi _{B}^{c_2}}^2+{\pi _{C}^{c_2}}^2}}=0\\
	\end{array}
	\label{eqn:init}
	\end{equation}
	
	where $d\left( S_{c_1},\pi _2 \right)$ and $d\left( E_{c_1},\pi _2 \right)$ are the distances from the two endpoints to the plane $\pi _2$. $S_{{c_1}}=\frac{1}{{{\lambda _s}}}(u_s^{{c_1}},v_s^{{c_1}},1)$ and $E_{{c_1}}=\frac{1}{{{\lambda _e}}}(u_e^{{c_1}},v_e^{{c_1}},1)$ are the spatial coordinates of the two endpoints. 
	So (\ref{eqn:init}) sets up two equations for two variables and the values of $\lambda _s$ and $\lambda _e$ can be calculated. 
	
	In reality, we often observe the same line in multiple frames, so we can get overdetermined equations(\ref{eqn:overdetermin}).
	\begin{equation}
	\left[ {\begin{array}{*{20}{c}}
		{\begin{array}{*{20}{c}}
			{d({S_{{c_1}}},{\pi _2})} \\ 
			{d({E_{{c_1}}},{\pi _2})} 
			\end{array}} \\ 
		\vdots  \\ 
		{\begin{array}{*{20}{c}}
			{d({S_{{c_1}}},{\pi _i})} \\ 
			{d({E_{{c_1}}},{\pi _i})} 
			\end{array}} 
		\end{array}} \right] = 0
	\label{eqn:overdetermin}
	\end{equation}
	We can use the nonlinear least squares method \cite{rosipal2011nonlinear} to optimize the solution $\lambda _s,\lambda _e$ from the equations (\ref{eqn:overdetermin}).

 \begin{prop}
	This proposed multi-frame optimization-based line triangulation approach can avoid the introduction of erroneous initialization results in the degraded scenes. 
 \end{prop}   
    For example, in a line degeneration scenario such as when the transformation between $c_1$ and $c_2$ contains only rotation, the Plücker matrix method\cite{hartley2003multiple} uses the following equation to conduct initialization.
	\begin{equation}
	\left[ {\begin{array}{*{20}{c}}
		{{{\left[ {{\mathbf d_{{c_1}}}} \right]}_\times}}&{{\mathbf n_{{c_1}}}} \\ 
		{ - \mathbf n_{c_1}^\top}&0 
		\end{array}} \right] = {\pi _1}{\pi _2^\top} - {\pi _2}{\pi _1^\top}
	\end{equation}
	where $\left[ \cdot \right]_\times$ denotes the skew-symmetric matrix of a 3D Vector. If the rotation-only case occurs, the wrong initialization value will be given to the direction vector $\mathbf d_{{c_1}}$\cite{hartley2003multiple}. But using the inverse depth initialization in (\ref{eqn:overdetermin}), in rotation-only case, $c_i$ equals to $c_1$ in $\pi_{c_1}$ which is $(0,0,0)$, so $\pi _{D}^{c_i}=0$. 
	At the same time the normal vectors of the two projection planes are parallel in $\pi_{c_{1}}$, which makes product of a plane normal vector and a point in the plane constantly equal to 0.
	So the distances $d\left( S_{c_1},\pi _i \right)$ and $d\left( E_{c_1},\pi _i \right)$ will always be 0. 
	We can detect such zero equals zero situation and avoid conducting initialization in the rotation-only case. This avoids the introduction of wrong initialization solutions. Also because of our multi-frame optimization approach, line feature initialization can be done as long as there are two frames which are not degraded. 

	
	\section{IDLS: Inverse Depth Line SLAM}
\begin{figure*}[!t]
		\centering
		\includegraphics[width=0.7\linewidth]{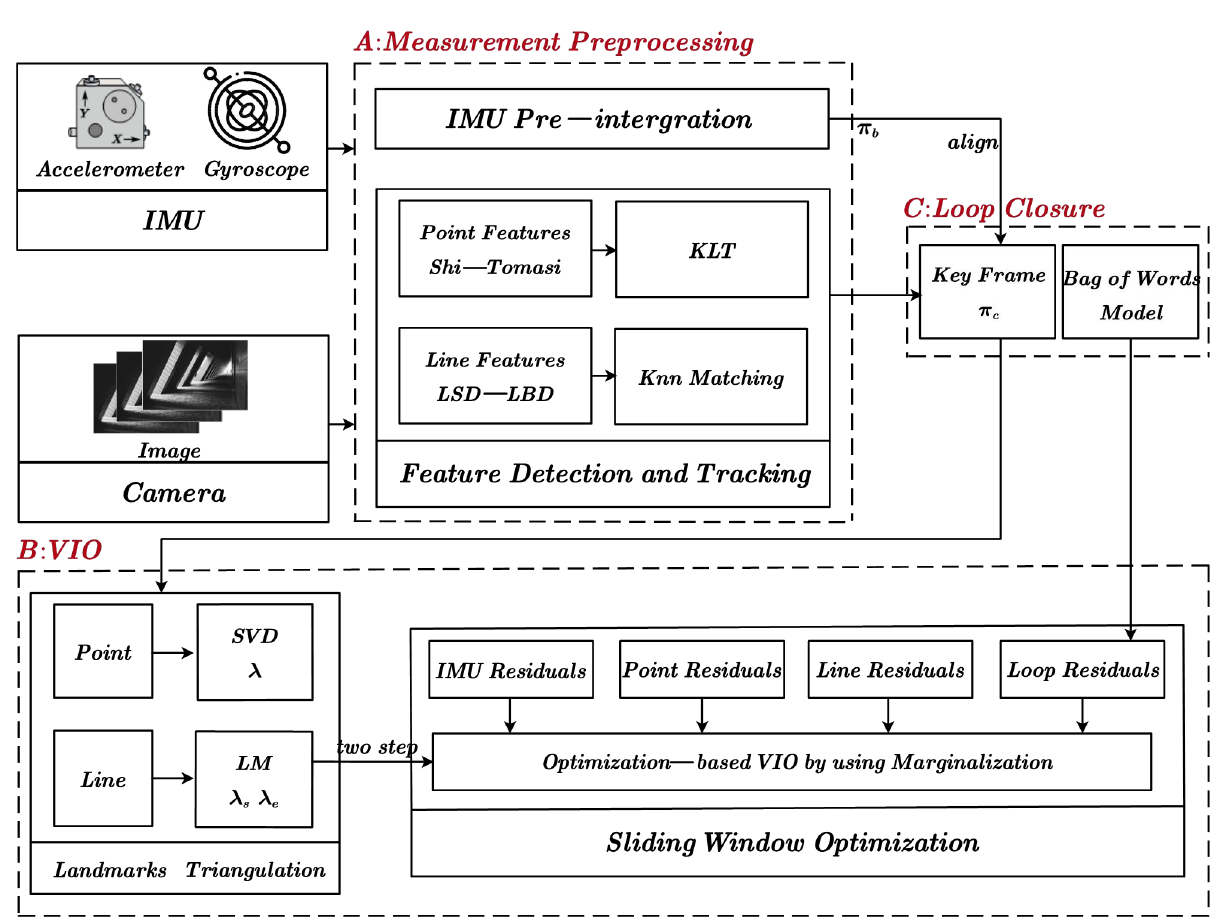}
		\caption{System overview of IDLS. It consists of three threads: measurement preprocessing, VIO, and loop closure.}
		\label{fig:framework}
	\end{figure*}

	Based on the inverse depth line representation, the residue error function, and the initialization method, 
	we integrate the inverse depth line features and develop IDLS. This section introduces the IDLS framework. The system architecture is shown in Fig.\ref{fig:framework}. IDLS contains three threads: (1) measurement processing, which conducts IMU pre-integration, point, and line feature extraction and matching; (2) the Vision Inertial Odometry (VIO) thread, which calculates the camera ego-motion and point-line locations by graph optimization in a sliding window; (3) the loop closure thread, which uses the bag-of-words model of points to conduct loop detection. 
	
	\subsection{Measurement Processing}
	
	The first IDLS thread extracts data from the camera's and IMU's raw measurements.
	For the images captured by the camera, the algorithm will detect and track point and line features.
	Point features are found using Shi-Tomasi\cite{shi1994good}, tracked using KLT\cite{baker2004lucas}, and inliers are found using RANSAC-based epipolar geometry constraint\cite{hartley2003multiple}. Regarding the line features, the LSD detector\cite{von2008lsd}  and LBD descriptor\cite{zhang2013efficient} are used to identify the line segments in the new frame.
	We match the line segments with LBD descriptors in two consecutive frames by KNN\cite{kaehler2016learning} to track them, and remove outliers of line features by geometric constraints. The IMU information between two adjacent image frames is pre-integrated to be aligned with the visual information. We pre-integrate new IMU measurements between two consecutive frames to update the newest body states.
	
	With these two types of preprocessed information, IDLS initializes some necessary parameters to trigger the visual-inertial odometry thread. First, it constructs a graph structure of poses, including the points and lines of the environment in the selected keyframes. The keyframes are selected to reduce the number of estimated states and the amount of optimization calculation. Here, we use the VINS-Mono\cite{qin2018vins} keyframe selection method. 
	There are two criteria for selecting the keyframes:
	The graph is then aligned with the pre-integrated IMU states using the initialization method in VINS-Mono\cite{qin2018vins}. The lastest body states are taken as the initial value in the sliding window optimization and the point-line triangulation.
	
	\subsection{Visual-Inertial Odometry}
	
	A two-step tightly-coupled optimization-based visual-inertial odometry (VIO) thread is used in IDLS, to make pose estimation more accuracy by minimizing all the measurement residuals. VIO mainly consists of three parts: \textbf{construction of sliding window, initialization of features, and optimization calculation of the states}. 
	The details of some specific steps are as followings:
	
	\subsubsection{Sliding Window Formulation}
	
	First, we adopt a fixed-size sliding window for achieving both computation efficiency and high accuracy. The definition of each state variable in the sliding window at time $i$ is:
	\begin{equation}
	\begin{gathered}
	\mathcal{X} = \left[ {{\mathbf x_0},{\mathbf x_1},...,{\mathbf x_n},\mathbf x_{bc},{\lambda _0},{\lambda _1},...,{\lambda _{{n_p}}},{\mathbf l_0},{\mathbf l_1},...,{\mathbf l_{{n_l}}}} \right] \hfill \\
	{\mathbf x_k} = \left[ {{p_{wb_k}},{q_{wb_k}},{v_{wb_k}},{b_a},{b_g}} \right],k \in \left[ {0,n} \right] \hfill \\
	{\mathbf x_{bc}} = \left[ {{p_{bc}},{q_{bc}}} \right] \hfill \\
	{\mathbf l_m} = \left[ {\lambda _{s_m},\lambda _{e_m}} \right],m \in \left[ {0,{n_l}} \right] \hfill \\ 
	\end{gathered}
	\label{eqn:state} 
	\end{equation}
	
	in which $\mathbf x_k$ consists of the $k$-th IMU body position ${p_{wb_k}}$, orientation ${q_{wb_k}}$, velocity ${v_{wb_k}}$ in $\pi _w$, acceleration bias $b_a$, gyroscope bias $b_g$. The total number of keyframes, space points and lines in the sliding window are respectively defined as $n$, $n_p$, and $n_l$. $\lambda$ represents the inverse distance of a point feature from its first observed keyframe. $\mathbf l$ is the inverse depth of a 3D line feature which is calculated from its first observed keyframe. 
	
	\begin{figure}[!t]
		\centering
		\includegraphics[width=1\linewidth]{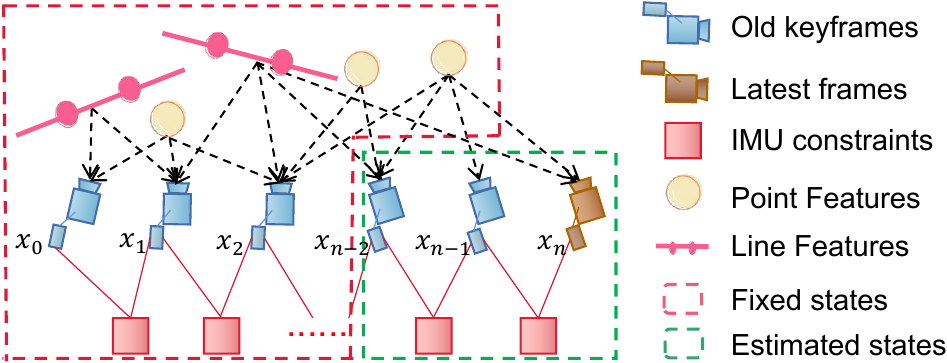}
		\caption{Factor graphs of visual–inertial SLAM with point and line landmarks.}
		\label{fig:slide-window}
	\end{figure}
	
	For vision-inertial fusion, factor-graph-based tightly-coupled framework is used to optimize each state variable as shown in Fig.\ref{fig:slide-window}. The nodes are the IMU body states to be optimized and the 3D features to be optimized. The edges represent the measurements from IMU pre-integration and the visual observations. The edges serve as constraints between the nodes. The continuous IMU body states are constrained by the IMU pre-integrated measurements, which also give initial values for bundle adjustment (the image is non-convex with multiple local optimal solutions). The visual measurements use bundle adjustment to constrain the camera poses and 3D map landmarks. Then, we use factor-graph-based method to optimize the body poses and 3D map landmarks by minimizing the re-projection error in image planes and the continuous IMU body states:  
	
	\begin{equation}
	\begin{array}{l}
	\mathop {min} \limits_{\mathcal{X}}\rho (\left\| r_p-J_p\mathcal{X} \right\| _{\sum{p}}^{2})+\sum\limits_{i\in B}{\rho (\left\| r_b(z_{b_ib_{i+1}},\mathcal{X} ) \right\| _{\sum{b_ib_{i+1}}}^{2})}
	\\
	+\sum\limits_{(i,j)\in F}{\rho (\left\| r_f(z_{_{f_j}}^{c_i},\mathcal{X} ) \right\| _{\sum_{f_j}^{c_i}}^{2})}+\sum\limits_{(i,j)\in L}{\rho (\left\| r_l(z_{_{l_j}}^{c_i},\mathcal{X} ) \right\| _{\sum_{l_j}^{c_i}}^{2})}
	\\
	+\sum\limits_{(i,j)\in Loop}{\rho (\left\| r_L(\hat{z}_{i}^{j},\mathcal{X} ,\hat{q}_{v}^{w},\hat{p}_{v}^{w}) \right\| _{\sum^j_i}^{2})}
	\\
	\end{array}
	\label{eqn:residual} 
	\end{equation}
	
	in which ${{r_b}({z_{{b_i}{b_{i + 1}}}},\mathcal{X})}$ is the IMU measurement residual between the body state $x_i$ and $x_{i+1}$. $B$ is the set of all IMU pre-integrated measurements within sliding window. ${{r_f}(z_{_{{f_j}}}^{{c_i}},\mathcal{X})}$ and ${{r_l}(z_{_{{l_j}}}^{{c_i}},\mathcal{X})}$ are the point feature re-projection residual and the line feature re-projection residual, respectively. $F$ and $L$ are the sets of point features and line features extracted from camera frames. ${{r_L}(\hat z_i^j,\mathcal{X},\hat q_v^w,\hat p_v^w)}$ is loop re-projection residual and $Loop$ represents the set of loop closure frames. After marginalizing a frame from the sliding window, $\{r_p,J_p\}$ is prior information that can be calculated \cite{shen2015tightly}, and the prior Jacobian matrix $J_p$ is from the Hessian matrix after the previous optimization. We can use $\rho $  (Cauchy robust function) to constraint outliers. Here the residual edges of VINS-Mono are changed from optimizing feature points to optimize the combination of point-line features.
	
	
	\subsubsection{Initialization Features in Sliding Window}
	
	In the sliding window, we reconstruct space points and lines by triangulating the 2D point and line feature correspondences of successive frames. The inverse depth is used to parameterize space points.\cite{civera2008inverse}, and the inverse depth of the two endpoints of the first observed frame in (\ref{eqn:overdetermin}) is used to parameterize space lines. 
	
	Among them, the initial camera poses of the $i$-th and $j$-th frames are obtained through IMU pre-integration, so that we can focus on optimizing the inverse depth values of the feature points and feature lines.  Giving the initial values, a two step graph optimization scheme is exploited for quick convergence.

	\subsubsection{Two-step Graph Optimization}
    \label{sec:two-step}
 
\begin{figure}[!t]
		\centering
		\includegraphics[width=1\linewidth]{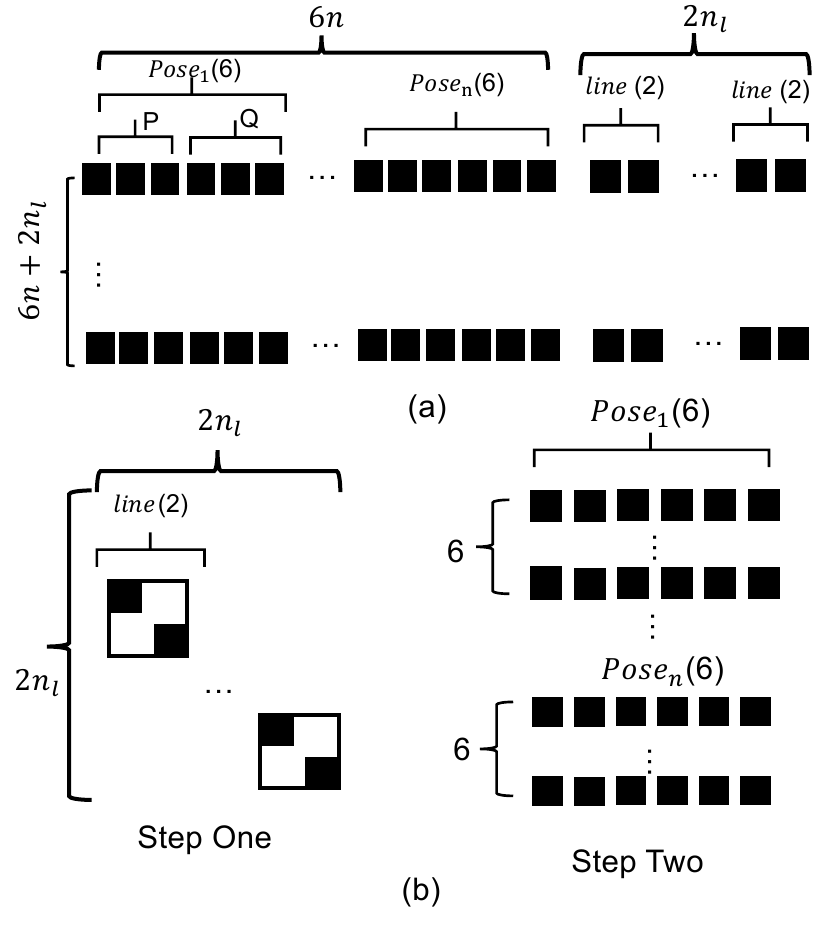}	
		\caption{Hessian matrix of line residual in traditional optimization (a) and our two-step optimization (b). Assuming we have a total of $n_l$ lines within $n$ key frames.  the complexity is $O((2\times n_l +2\times n)^2)$ using the Hessian matrix in (a). However, if we use the two-step optimization in (b), the complexity is $O((2\times n_l)^2+n\times6^2)$. The complexity is much reduced.}
		
		\label{fig:jacobi}
		
	\end{figure}

	Levenberg–Marquard algorithm is used to solve the nonlinear optimization problem(\ref{eqn:residual}). Then we can get the optimal state estimates $\mathcal X$ by iteratively updating the variables from an initial guess $\mathcal X_0$ as:
	
	\begin{equation}
	\mathcal X_{t + 1}^{'} = {\mathcal X_t} \oplus \delta \mathcal X
	\end{equation}
	
	in which $\oplus$ is the operator used to update parameters with increment $\delta \mathcal X$. At each iteration, we calculate the increment $\delta \mathcal X$ by Equation (\ref{eqn:update}):
	
	\begin{equation}
	({H_p} + {H_b} + {H_f} + {H_l} + {H_L})\delta \mathcal X = ({b_p} + {b_b} + {b_f} + {b_l} + {b_L})
 \label{eqn:update}
	\end{equation}
	
	in which ${H_p}$, ${H_b}$, ${H_f}$, ${H_l}$ and ${H_L}$ represent the Hessian matrices for prior residuals, IMU measurement residuals, point and line re-projection residuals ,and loop residuals, respectively. For residual $r_{(\cdot)}$, we have ${H_{\left( \cdot \right)}} = {J_{\left( \cdot \right)}^\top}\Sigma _{\left( \cdot \right)}^{ - 1}{J_{\left( \cdot \right)}}$ and ${b_{\left( \cdot \right)}} = -{J_{\left( \cdot \right)}^\top}\Sigma _{\left( \cdot \right)}^{ - 1}{r_{\left( \cdot \right)}}$, where $J_{\left( \cdot \right)}$ is the Jacobian matrix of residuals vector $r_{(\cdot)}$ with respect to $\delta \mathcal X$, and $\Sigma _{\left( \cdot \right)}$ is the covariance matrix of measurements. 
	Note that adding the line features in the cost function (\ref{eqn:residual}) enlarges the number of unknown variables. 
	Fig.\ref{fig:jacobi}(a) shows the correlation relationships among the states' and line parameters Hessian matrix. 
	
	\textbf{To make the optimization more efficient, we introduce a two-step approach to minimize the residual error in (\ref{eqn:residual})}. We first focus on the line re-projection residual, i.e., $\sum_{(i,j)\in L}{\rho (\left\| r_l(z_{_{l_j}}^{c_i},\mathcal{X} ) \right\| _{\sum^{c_i}_{l_j}}^{2})}$. 
	The camera postures and inverse depth of the line's endpoints are optimized using this residual form. 
	We divide the problem into two steps: (1) optimizing inverse depths of the line endpoints; (2) optimizing the camera poses. 
	The first step is analogous to the initialization phase, which is to determine  3D lines when the camera poses are known. As a result, we can lower the dimension of the Hessian matrix in each step as shown in Fig.~\ref{fig:jacobi}(b). 
	A two-step optimization algorithm is presented in Algorithm~\ref{alg1}. We first fix the camera poses using the predicted poses by IMU and use the least square optimization to obtain the line features'  endpoint inverse depths in (\ref{eqn:residual}). 
	It solves a linear equation array, which is efficient. 
	Then, in the second step, the two endpoint inverse depths of the line features are fixed and the camera poses and external parameters are optimized. Levenberg–Marquard algorithm is used to solve this dimension-reduced nonlinear optimization problem\cite{bloesch2015robust}. 
	
	
	\begin{algorithm}[H]
		\caption{Two-step minimization for $r_l$.}\label{alg:alg1}
		\begin{algorithmic}
			\WHILE{not converge}
			\STATE 1) Use the latest poses to fit 3D lines by $\hat{\mathbb{L}} = \{ \hat L|\hat L = \arg {\min _L}(d\left( S_{c_i},\pi _j \right),d\left( E_{c_i},\pi _j \right)),L \in \mathbb{L}\} $;
			\STATE 2) Fix $\hat{\mathbb{L}}$ and conduct one Levenberg-Marquardt step to update poses to reduce the cost $r_l$;
			\ENDWHILE		
		\end{algorithmic}
		\label{alg1}
	\end{algorithm}
	\vspace{-0.3in}
	
	%

	\subsection{Loop Closure}
	
	IDLS initiates the loop closure thread whenever the current frame is selected as a keyframe. This thread uses the original loop closure method of VINS-Mono to search for and identify instances of loop closure.\cite{qin2018vins, qin2018relocalization}. It uses the bag-of-words model\cite{zhang2010understanding} constructed by the point feature descriptors to detect loop closure and use PnP RANSAC\cite{lepetit2009epnp} to eliminate outliers. If loop closure occurs, relocation residual edge is added to the tightly-coupled optimization-based VIO.
	
	
	\section{Experiments}
	
	We use the public benchmark dataset, i.e., the EuRoc MAV Dataset\cite{burri2016euroc} and own dataset collected in challenging environments to evaluate the proposed methods' performances. We compared IDLS with four state-of-the-art monocular VIO algorithms:  VINS-Mono\cite{qin2018vins}(Point-based methods), PL-VINS\cite{fu2020pl}, PL-VIO\cite{he2018pl} (Point and line-based methods), and UV-SLAM\cite{lim2022uv} (Point and line-based method with vanishing point). All of the experiments were performed on a computer with an Intel Core i7-11700 with 2.50GHz, and implemented on Ubuntu 18.04 with ROS Melodic. 
	The line feature map and the motion trajectories of the UAV generated by IDLS on the MH-04-difficult sequence are shown in Fig.\ref{fig_first_case}.
	
	%
	
	\subsection{EuRoc MAV Dataset}
	
	The EuRoc MAV dataset contains two scenarios. (1) Machine Hall, which contains a variety of machinery and structures, as well as different types of textures and lighting conditions. (2) Vicon Room, which is an indoor environment equipped with a high precision Vicon motion capture system. We mainly use their medium and difficult trajectories.
	
	\subsubsection{\textbf{Accuracy Performance}}
	
	The absolute trajectory error (ATE) and the relative pose error (RPE) are used test the localization accuracy of different methods. 
	
	\textbf{ATE comparison}: The medium and difficult sequences in EuRoc datasets which include totally 7 sequences are tested. Table \ref{tab:ATE} shows the results of root mean square error (RMSE) of different algorithms. The results mainly evaluate the matching accuracy of frame to map.  Table \ref{tab:ATE} shows that:	
	\begin{table*}[]
		\normalsize
		\caption{RMSE ATE [M] Comparison of different methods}
		\centering
		\resizebox{\textwidth}{!}{	
			\begin{tabular}{|c|ccccc|cccc|}
				\hline
				\multirow{2}{*}{Datasets} & \multicolumn{5}{c|}{w/o loop}                                                                                & \multicolumn{4}{c|}{w/ loop}                                                   \\ \cline{2-10} 
				& \multicolumn{1}{c|}{VINS-Mono} & \multicolumn{1}{c|}{PL-VIO} & \multicolumn{1}{c|}{PL-VINS} &\multicolumn{1}{c|}{UV-SLAM}& IDLS           & \multicolumn{1}{c|}{VINS-Mono} & \multicolumn{1}{c|}{PL-VINS} &\multicolumn{1}{c|}{UV-SLAM}& IDLS           \\ \hline
				MH-03-Medium              & \multicolumn{1}{c|}{0.236}     & \multicolumn{1}{c|}{0.263}  & \multicolumn{1}{c|}{0.231}   &\multicolumn{1}{c|}{0.296}   & \textbf{0.204}\color{red}$( \downarrow 11.7\% )$ & \multicolumn{1}{c|}{0.104}     & \multicolumn{1}{c|}{0.099}& \multicolumn{1}{c|}{0.078}   & \textbf{0.071}\color{red}$( \downarrow 28.3\% )$ \\ \hline
				MH-04-Difficult           & \multicolumn{1}{c|}{0.376}     & \multicolumn{1}{c|}{0.360}  & \multicolumn{1}{c|}{0.282}   & \multicolumn{1}{c|}{0.294} &\textbf{0.270}\color{red}$( \downarrow 4.26\% )$ & \multicolumn{1}{c|}{0.220}     & \multicolumn{1}{c|}{0.202}   &\multicolumn{1}{c|}{0.179}& \textbf{0.170}\color{red}$( \downarrow 15.8\% )$ \\ \hline
				MH-05-Difficult           & \multicolumn{1}{c|}{0.295}     & \multicolumn{1}{c|}{0.279}  & \multicolumn{1}{c|}{0.272}   & \multicolumn{1}{c|}{0.275}&\textbf{0.239}\color{red}$( \downarrow 12.1\% )$ & \multicolumn{1}{c|}{0.242}     & \multicolumn{1}{c|}{0.226}   &\multicolumn{1}{c|}{0.145}& \textbf{0.137}\color{red}$( \downarrow 39.4\% )$ \\ \hline
				V1-02-Medium              & \multicolumn{1}{c|}{0.169}     & \multicolumn{1}{c|}{Fail}   & \multicolumn{1}{c|}{0.123}   &
				\multicolumn{1}{c|}{0.116}   & \textbf{0.112}\color{red}$( \downarrow 8.94\% )$ & \multicolumn{1}{c|}{0.091}     & \multicolumn{1}{c|}{0.079}   &\multicolumn{1}{c|}{0.079}   & \textbf{0.073}\color{red}$( \downarrow 7.59\% )$ \\ \hline
				V1-03-Difficult           & \multicolumn{1}{c|}{0.251}     & \multicolumn{1}{c|}{0.187}  & \multicolumn{1}{c|}{0.182} & \multicolumn{1}{c|}{0.228}    & \textbf{0.170}\color{red}$( \downarrow 6.59\% )$ & \multicolumn{1}{c|}{0.225}     & \multicolumn{1}{c|}{0.180}  & \multicolumn{1}{c|}{0.175} & \textbf{0.141}\color{red}$( \downarrow 21.7\% )$ \\ \hline
				V2-02-Medium              & \multicolumn{1}{c|}{0.166}     & \multicolumn{1}{c|}{0.156}  & \multicolumn{1}{c|}{0.151}   & \multicolumn{1}{c|}{0.148}& \textbf{0.113}\color{red}$( \downarrow 25.2\% )$ & \multicolumn{1}{c|}{0.139}     & \multicolumn{1}{c|}{0.133}   & \multicolumn{1}{c|}{0.115}& \textbf{0.074}\color{red}$( \downarrow 44.4\% )$ \\ \hline
				V2-03-Difficult           & \multicolumn{1}{c|}{0.297}     & \multicolumn{1}{c|}{0.270}  & \multicolumn{1}{c|}{0.237}   & \multicolumn{1}{c|}{0.239}& \textbf{0.225}\color{red}$( \downarrow 5.60\% )$ & \multicolumn{1}{c|}{0.215}     & \multicolumn{1}{c|}{0.196}   & \multicolumn{1}{c|}{0.160}& \textbf{0.143}\color{red}$( \downarrow 27.0\% )$ \\ \hline
			\end{tabular}
		}
		\label{tab:ATE}  
	\end{table*}

	\begin{itemize}
		\item IDLS yields better localization accuracy in medium and difficult environments. For example, the ATE of our method is much less than that of VINS-Mono on the MH-04-difficult sequence. This shows that the point-line combination approach can indeed improve SLAM accuracy. Also our method is $4.3\%$ to $25.2\%$  more accurate than PL-VINS, which shows that the inverse depth representation of line features indeed provide better accuracy than the orthonormal representation.  IDLS also performs better than PL-VIO, UV-SLAM  in both settings of without or with loop closure.  
		\item Loop closure (loop) is very effective to eliminate accumulative error and is applicable to all seven sequences. Using IDLS as an illustration, loop closure reduces the MH-04 challenging sequence's ATE from 0.270 to 0.170. 
	\end{itemize}
	
	\textbf{RPE comparison}: 
	In comparing relative pose errors, we also test the medium and difficult sequences in EuRoc dataset. Table~\ref{tab:rpe1} and Table~\ref{tab:rpe2} show the results of different algorithms. RPE includes two parts: translational error and rotational error. 
	 Table~\ref{tab:rpe1} and Table~\ref{tab:rpe2} show that:
		IDLS outperforms the other four methods in terms of the RPE performance.  UV-SLAM has close performances in rotational errors with the addition of vanishing point optimisation.  In  degradated scenes, IDLS ensures  the wrong line direction vectors are not introduced in trajectory optimization. This gives us a lower trajectory drift, but the rotation angle calculation is sometimes not as good as UV-SLAM with the addition of vanishing point detection for line degradation environments.

	
	\begin{table*}[]
		\normalsize
		\caption{RMSE RPE [M] or [RAD] Comparison of different methods Without Loop}
		\centering
		\resizebox{\textwidth}{!}{		
			\begin{tabular}{|c|llllllllll|}
				\hline
				\multirow{3}{*}{Datasets} & \multicolumn{10}{c|}{w/o loop}                                                                                                                                                                                                                                                                             \\ \cline{2-11} 
				& \multicolumn{2}{c|}{VINS-Mono}                                & \multicolumn{2}{c|}{PL-VIO}                                    &\multicolumn{2}{c|}{PL-VINS}                                    &\multicolumn{2}{c|}{UV-SLAM}       &\multicolumn{2}{c|}{IDLS}                                                       \\ \cline{2-11} 
				& \multicolumn{1}{c|}{Trans.}   & \multicolumn{1}{c|}{Rot.}     & \multicolumn{1}{c|}{Trans.}           & \multicolumn{1}{c|}{Rot.}              & \multicolumn{1}{c|}{Trans.}   & \multicolumn{1}{c|}{Rot.}              & \multicolumn{1}{c|}{Trans.}            & \multicolumn{1}{c|}{Rot.}&
				\multicolumn{1}{c|}{Trans.}            & \multicolumn{1}{c|}{Rot.}              \\ \hline
				MH-03-Medium              & \multicolumn{1}{l|}{0.01174}  & \multicolumn{1}{l|}{0.00206}  & \multicolumn{1}{l|}{0.006935}         & \multicolumn{1}{l|}{0.000456}          & \multicolumn{1}{l|}{0.006569} & \multicolumn{1}{l|}{0.000433} &
				\multicolumn{1}{l|}{0.007906}          & \multicolumn{1}{l|}{\textbf{0.00037}} & \multicolumn{1}{l|}{\textbf{0.006388}} & 0.000471                               \\ \hline
				MH-04-Difficult           & \multicolumn{1}{l|}{0.01258}  & \multicolumn{1}{l|}{0.001431} & \multicolumn{1}{l|}{0.011454}         & \multicolumn{1}{l|}{0.001274}          & \multicolumn{1}{l|}{0.008322} & \multicolumn{1}{l|}{0.000871}          &
				\multicolumn{1}{l|}{0.008814} & \multicolumn{1}{l|}{0.000556}          & \multicolumn{1}{l|}{\textbf{0.00722}}  & \textbf{0.000456}                      \\ \hline
				MH-05-Difficult           & \multicolumn{1}{l|}{0.01145}  & \multicolumn{1}{l|}{0.001048} & \multicolumn{1}{l|}{0.007189}         & \multicolumn{1}{l|}{0.000544}          & \multicolumn{1}{l|}{0.007174} & \multicolumn{1}{l|}{0.000321}          &
				\multicolumn{1}{l|}{0.007571} & \multicolumn{1}{l|}{0.0004}&
				\multicolumn{1}{l|}{\textbf{0.006675}} & \textbf{0.000333}                      \\ \hline
				V1-02-Medium              & \multicolumn{1}{l|}{0.012745} & \multicolumn{1}{l|}{0.00278}  & \multicolumn{1}{l|}{Fail}             & \multicolumn{1}{l|}{Fail}              & \multicolumn{1}{l|}{0.01197}  & \multicolumn{1}{l|}{0.001598}          &
				\multicolumn{1}{l|}{0.006563}  & \multicolumn{1}{l|}{\textbf{0.000551}}          &
				\multicolumn{1}{l|}{\textbf{0.005159}} & 0.001928                     \\ \hline
				V1-03-Difficult           & \multicolumn{1}{l|}{0.126693} & \multicolumn{1}{l|}{0.00193}  & \multicolumn{1}{l|}{\textbf{0.00597}} & \multicolumn{1}{l|}{0.002394} & \multicolumn{1}{l|}{0.007986} & \multicolumn{1}{l|}{0.006758}          &
				\multicolumn{1}{l|}{0.00698} & \multicolumn{1}{l|}{\textbf{0.001939}}          & \multicolumn{1}{l|}{0.006186}          & 0.003094                               \\ \hline
				V2-02-Medium              & \multicolumn{1}{l|}{0.013071} & \multicolumn{1}{l|}{0.002729} & \multicolumn{1}{l|}{0.115221}         & \multicolumn{1}{l|}{0.002006}          & \multicolumn{1}{l|}{0.004547}&
				\multicolumn{1}{l|}{\textbf{0.000658}} &
				\multicolumn{1}{l|}{0.006291}&
				\multicolumn{1}{l|}{0.000838}& \multicolumn{1}{l|}{\textbf{0.004209}} & 0.000938                               \\ \hline
				V2-03-Difficult           & \multicolumn{1}{c|}{0.126455} & \multicolumn{1}{c|}{0.002271} & \multicolumn{1}{c|}{0.012743}         & \multicolumn{1}{c|}{0.0054}            & \multicolumn{1}{c|}{0.006894} & \multicolumn{1}{c|}{0.002344}          &
				\multicolumn{1}{c|}{0.008507} & \multicolumn{1}{c|}{0.002347}          & \multicolumn{1}{c|}{\textbf{0.006122}} & \multicolumn{1}{c|}{\textbf{0.002287}} \\ \hline
			\end{tabular}
		}
		\label{tab:rpe1}
	\end{table*}
	
	\begin{table*}[]
		\normalsize
		\caption{RMSE RPE [M] or [RAD] Comparison of different methods With Loop}
		\centering		
		\begin{tabular}{|c|llllllll|}
			\hline
			\multirow{3}{*}{Datasets} & \multicolumn{8}{c|}{w/ loop}                                                                                                                                                                                    \\ \cline{2-9} 
			& \multicolumn{2}{c|}{VINS-Mono}                                         & \multicolumn{2}{c|}{PL-VINS}& 
			\multicolumn{2}{c|}{UV-SLAM} & \multicolumn{2}{c|}{IDLS}                                              \\ \cline{2-9} 
			& \multicolumn{1}{c|}{Trans.}   & \multicolumn{1}{c|}{Rot.}              & \multicolumn{1}{c|}{Trans.}   & \multicolumn{1}{c|}{Rot.}     & \multicolumn{1}{c|}{Trans.}            & \multicolumn{1}{c|}{Rot.} &\multicolumn{1}{c|}{Trans.}            & \multicolumn{1}{c|}{Rot.}     \\ \hline
			MH-03-Medium              & \multicolumn{1}{l|}{0.013592} & \multicolumn{1}{l|}{0.002459}          & \multicolumn{1}{l|}{0.010582} & \multicolumn{1}{l|}{0.001157} &
			\multicolumn{1}{l|}{0.010381} & \multicolumn{1}{l|}{0.000469} &
			\multicolumn{1}{l|}{\textbf{0.009249}} & \textbf{0.000678}             \\ \hline
			MH-04-Difficult           & \multicolumn{1}{l|}{0.01521}  & \multicolumn{1}{l|}{0.001806}          & \multicolumn{1}{l|}{0.017375} & \multicolumn{1}{l|}{0.002498} &
			\multicolumn{1}{l|}{0.010926} & \multicolumn{1}{l|}{0.000962} & \multicolumn{1}{l|}{\textbf{0.009842}} & \textbf{0.000954}             \\ \hline
			MH-05-Difficult           & \multicolumn{1}{l|}{0.01601}  & \multicolumn{1}{l|}{\textbf{0.000823}} & \multicolumn{1}{l|}{0.020025} & \multicolumn{1}{l|}{0.002436} &
			\multicolumn{1}{l|}{0.015899} & \multicolumn{1}{l|}{0.001871} & \multicolumn{1}{l|}{\textbf{0.014695}} & 0.001394                      \\ \hline
			V1-02-Medium              & \multicolumn{1}{l|}{0.010025} & \multicolumn{1}{l|}{0.002839}          & \multicolumn{1}{l|}{0.011136} & \multicolumn{1}{l|}{0.004869} &
			\multicolumn{1}{l|}{\textbf{0.006745}} & \multicolumn{1}{l|}{\textbf{0.000562}} & \multicolumn{1}{l|}{0.009248} & 0.001629             \\ \hline
			V1-03-Difficult           & \multicolumn{1}{l|}{0.134445} & \multicolumn{1}{l|}{0.002084} & \multicolumn{1}{l|}{0.009435} & \multicolumn{1}{l|}{0.003023} &
			\multicolumn{1}{l|}{\textbf{0.007323}} & \multicolumn{1}{l|}{\textbf{0.002080}} & \multicolumn{1}{l|}{0.007862} & 0.007036                      \\ \hline
			V2-02-Medium              & \multicolumn{1}{l|}{0.125121} & \multicolumn{1}{l|}{0.002058}          & \multicolumn{1}{l|}{0.006793} & \multicolumn{1}{l|}{0.003306} &
			\multicolumn{1}{l|}{0.006273} & \multicolumn{1}{l|}{\textbf{0.000867}} & \multicolumn{1}{l|}{\textbf{0.005904}} & 0.002136             \\ \hline
			V2-03-Difficult           & \multicolumn{1}{c|}{0.132729} & \multicolumn{1}{c|}{\textbf{0.002382}} & \multicolumn{1}{c|}{0.011246} & \multicolumn{1}{c|}{0.005627} &
			\multicolumn{1}{c|}{\textbf{0.008469}} & \multicolumn{1}{c|}{0.002417}& \multicolumn{1}{c|}{0.010067} & \multicolumn{1}{c|}{0.004776} 
			\\ \hline
		\end{tabular}
		\label{tab:rpe2}
	\end{table*}
	
\begin{figure*}[h]
	\centering
	\vspace{-0.6in}
	{~~~~~~~~~MH-03-Medium~~~~~~~~~}
	{~~~~~~~~~MH-05-Difficult~~~~~~~~~}
	{~~~~~~~~~V1-02-Medium~~~~~~~~~}
	{~~~~~~~~~V2-03-Difficult~~~~~~~~~}
	
	\subfloat[]{\rotatebox{90}{~~~~~~~~~~VINS-MONO}\includegraphics[width=0.25\linewidth]{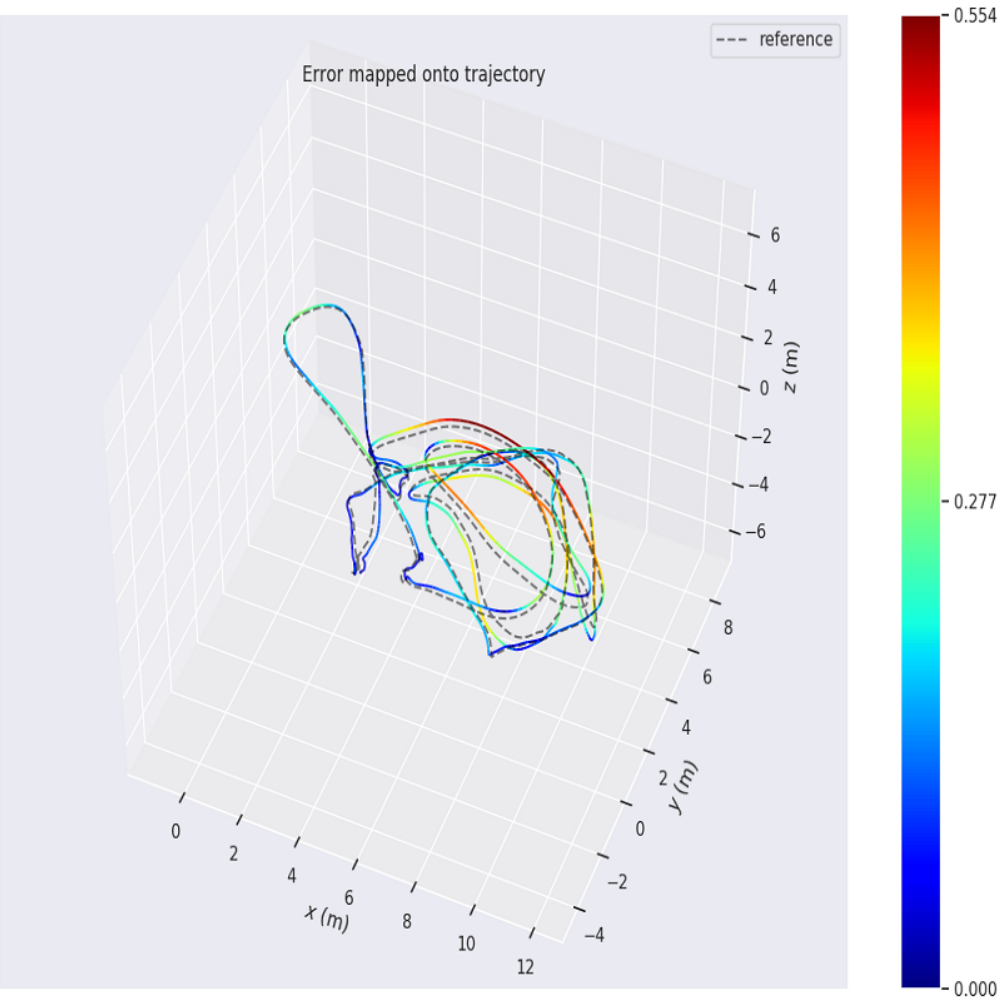}}
	\subfloat[]{\includegraphics[width=0.25\linewidth]{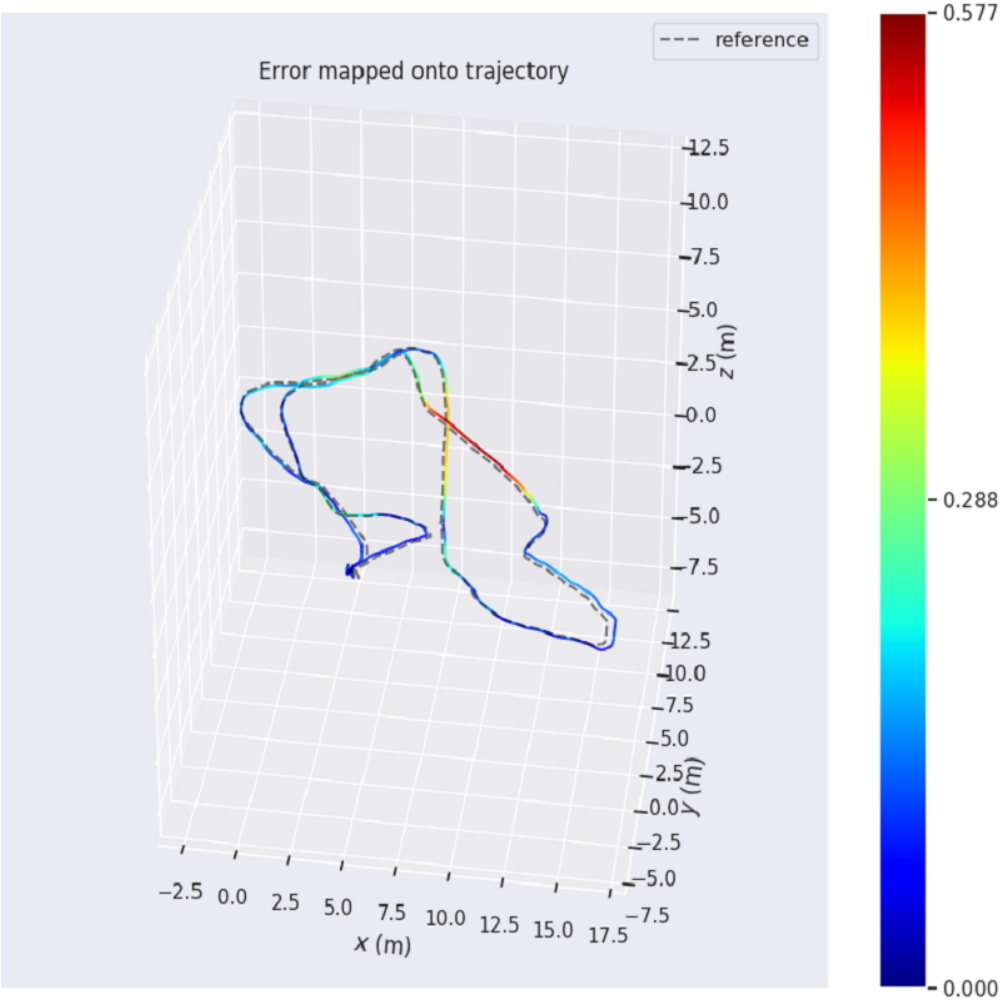}}
	\subfloat[]{\includegraphics[width=0.25\linewidth]{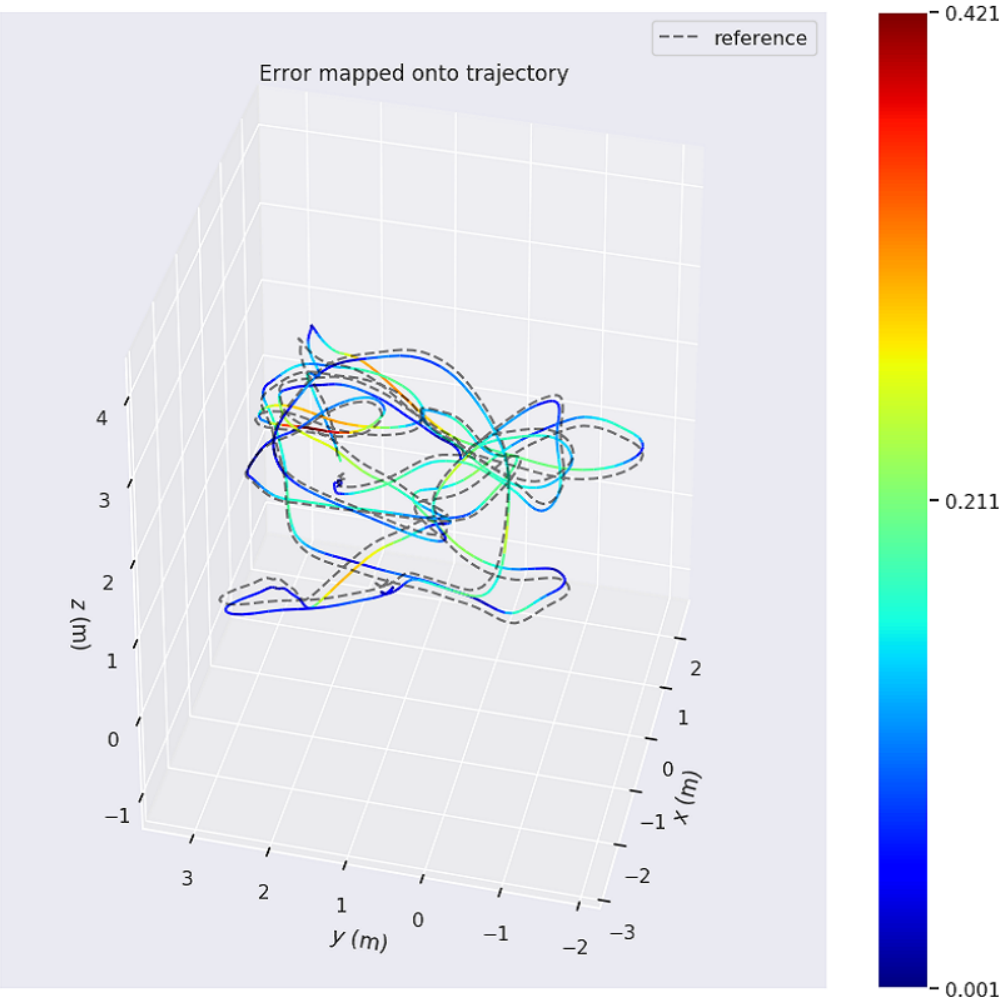}}
	\subfloat[]{\includegraphics[width=0.25\linewidth]{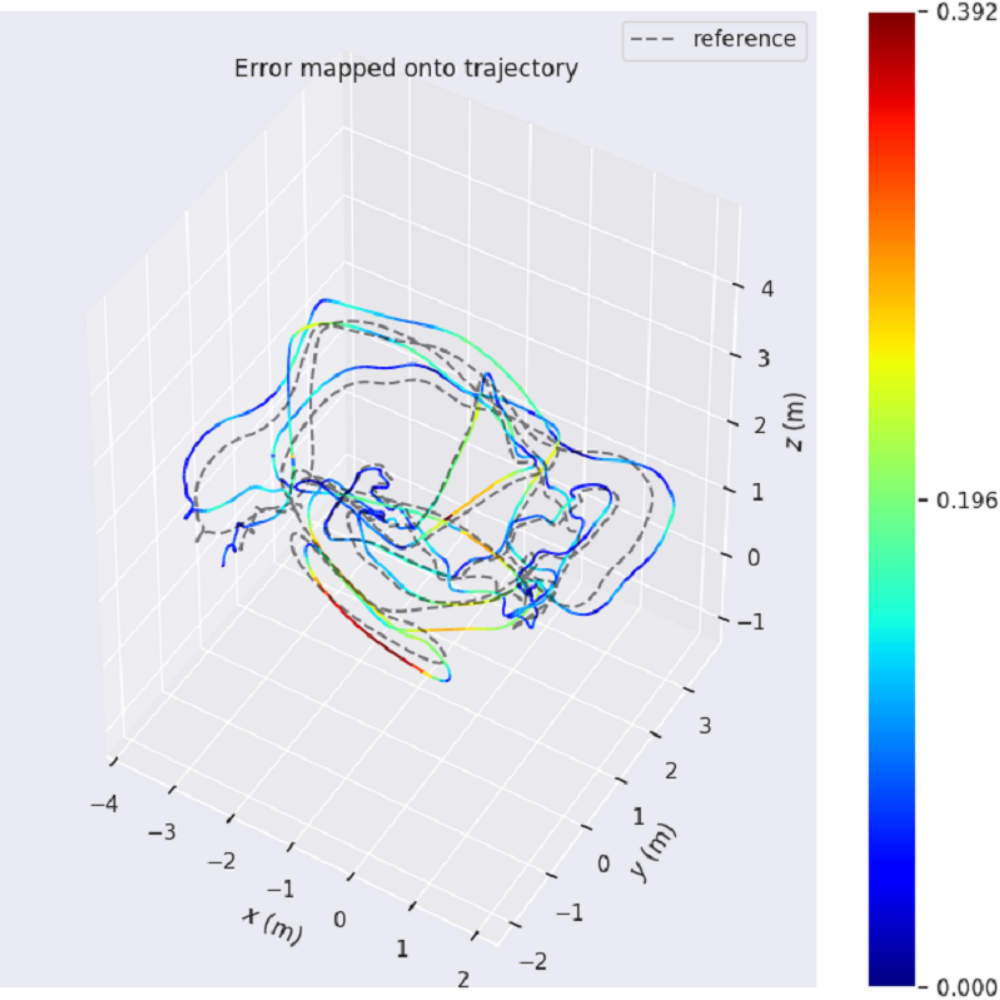}}
	\vspace{-0.1in}
	\subfloat[]{\rotatebox{90}{~~~~~~~~~~~~PL-VIO}\includegraphics[width=0.25\linewidth]{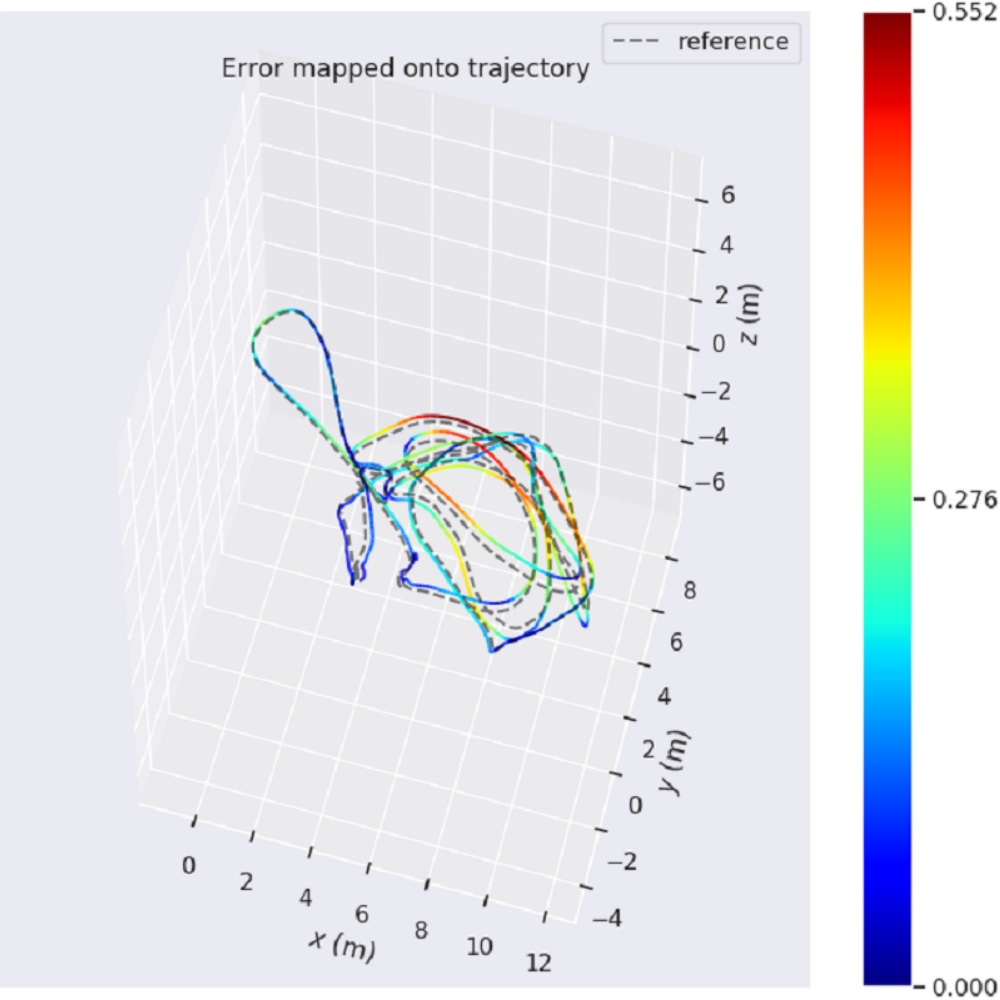}}
	\subfloat[]{\includegraphics[width=0.25\linewidth]{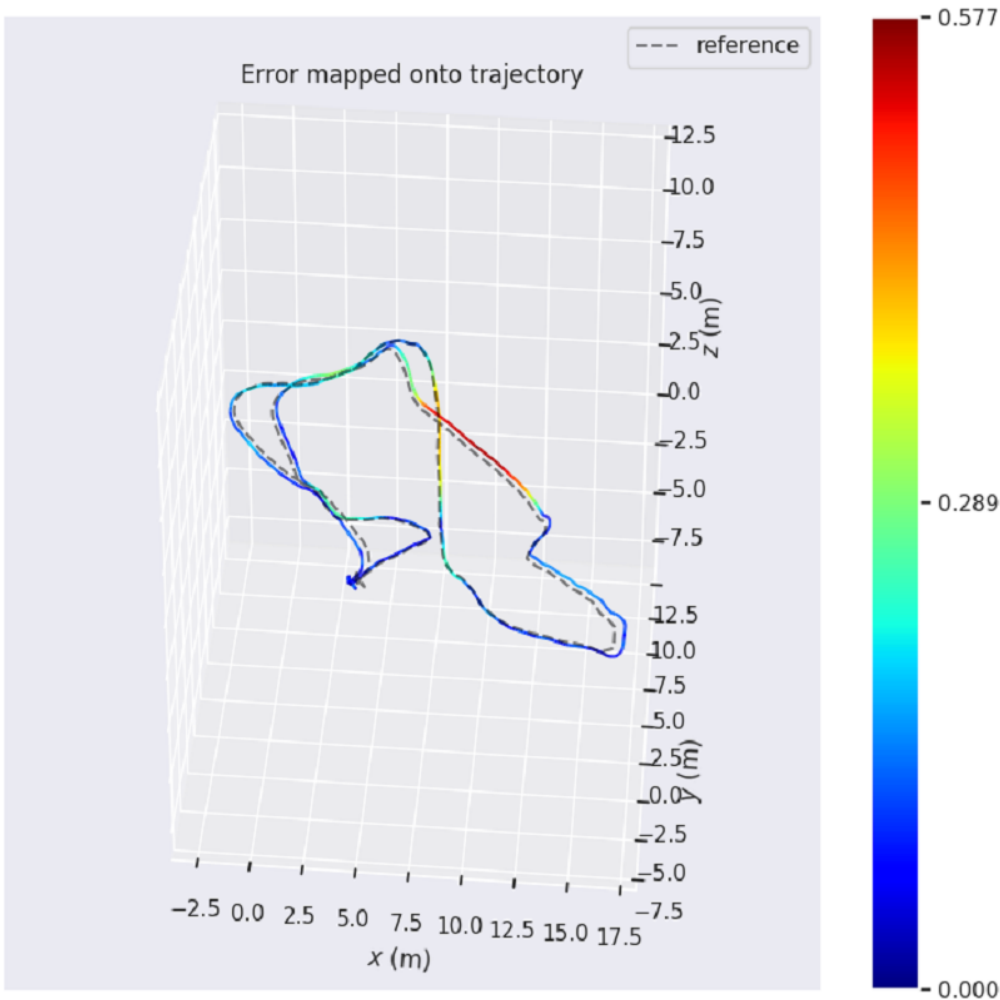}}
	\subfloat[]{\includegraphics[width=0.25\linewidth]{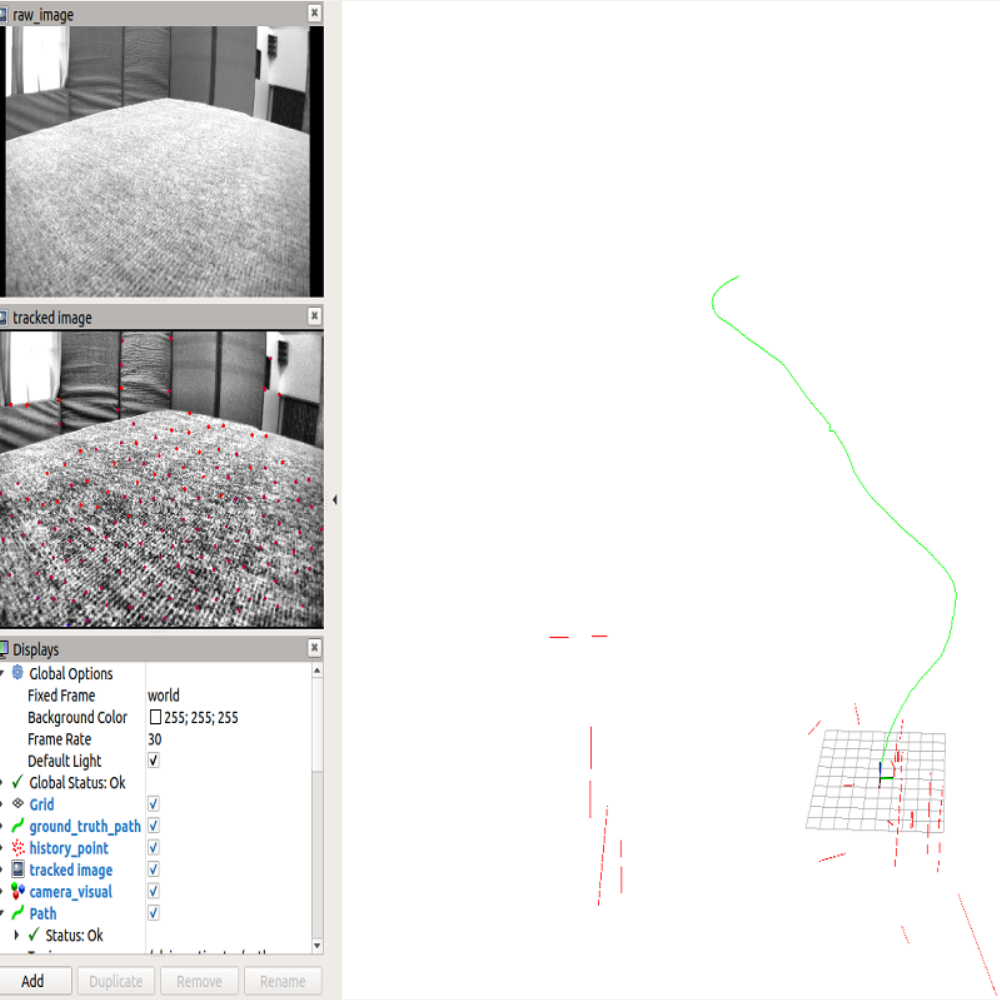}}
	\subfloat[]{\includegraphics[width=0.25\linewidth]{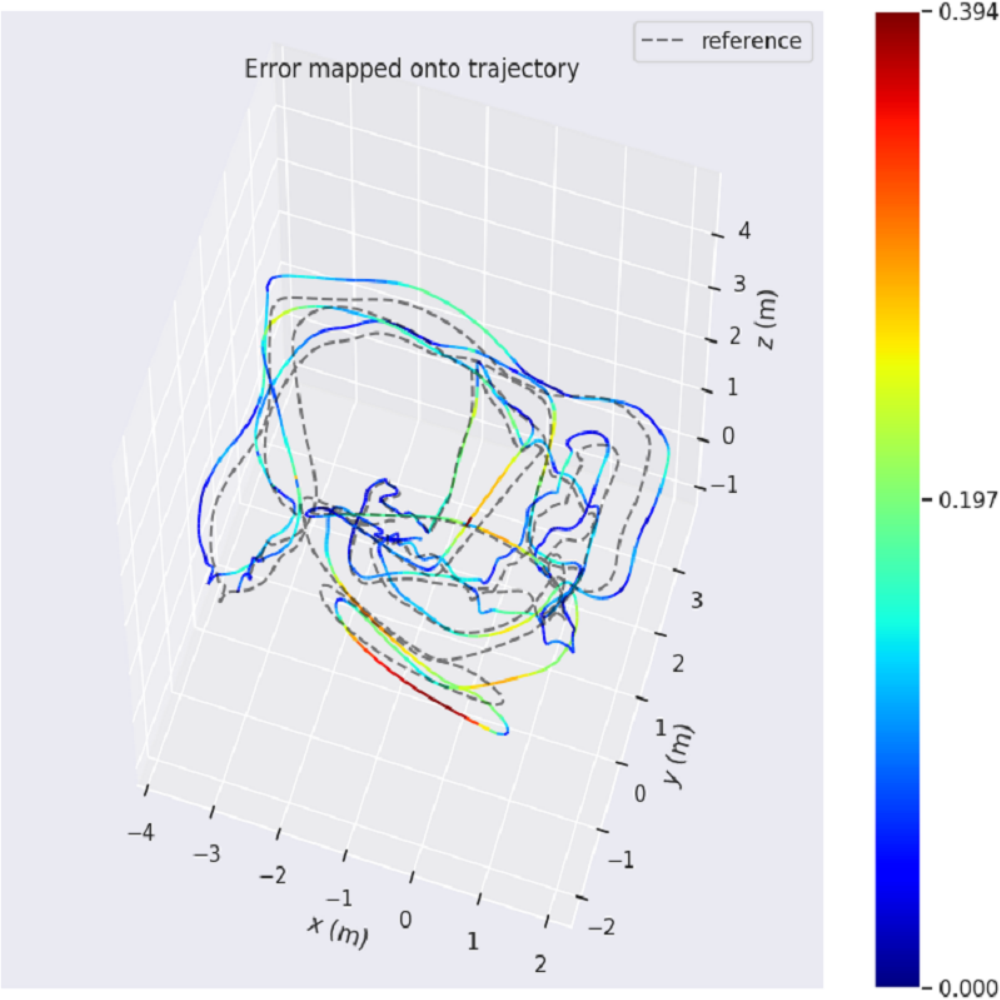}}
	\vspace{-0.1in}
	\subfloat[]{\rotatebox{90}{~~~~~~~~~~~~PL-VINS}\includegraphics[width=0.25\linewidth]{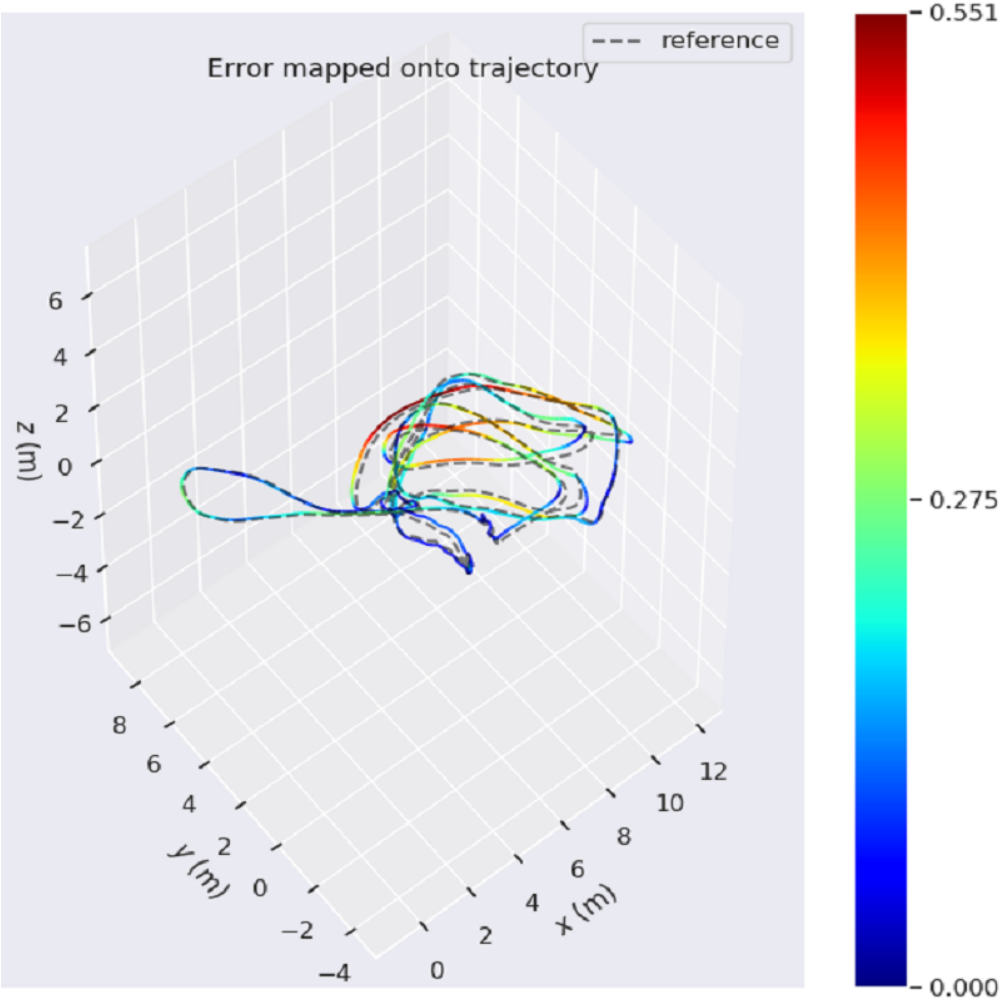}}
	\subfloat[]{\includegraphics[width=0.25\linewidth]{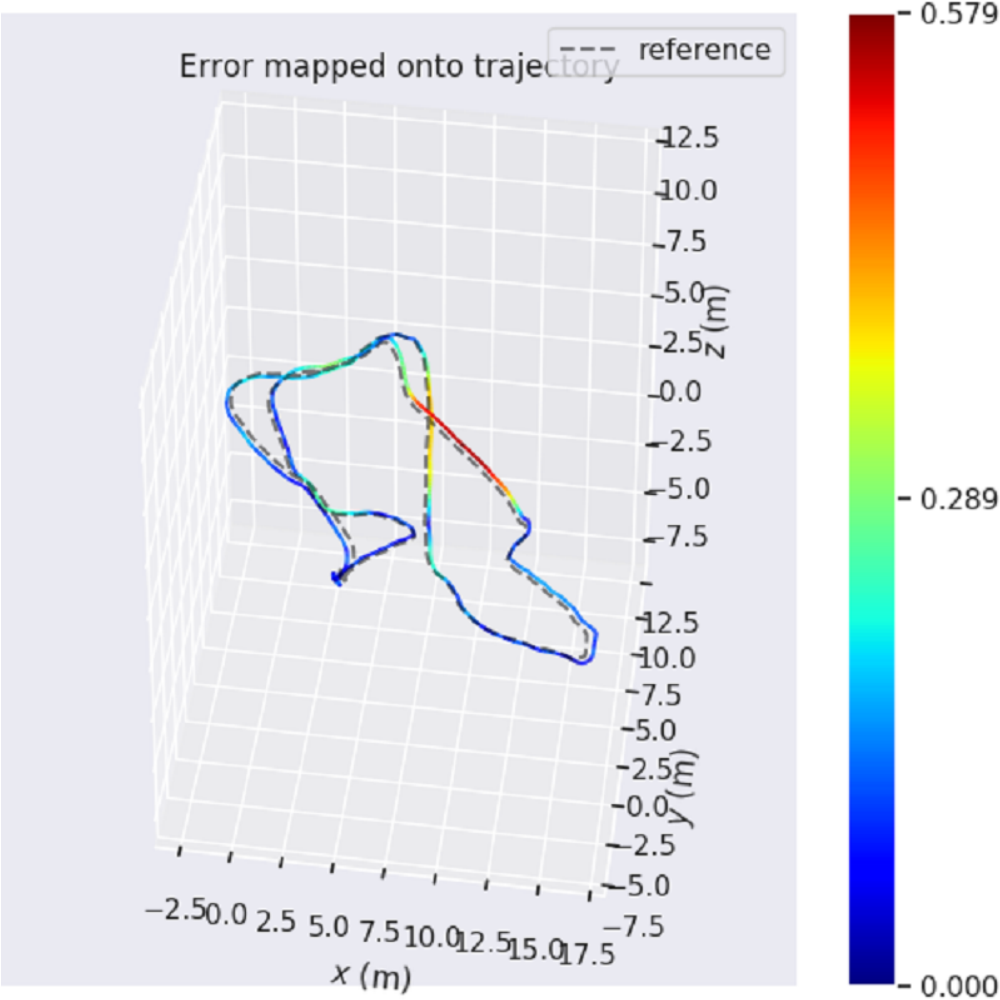}}
	\subfloat[]{\includegraphics[width=0.25\linewidth]{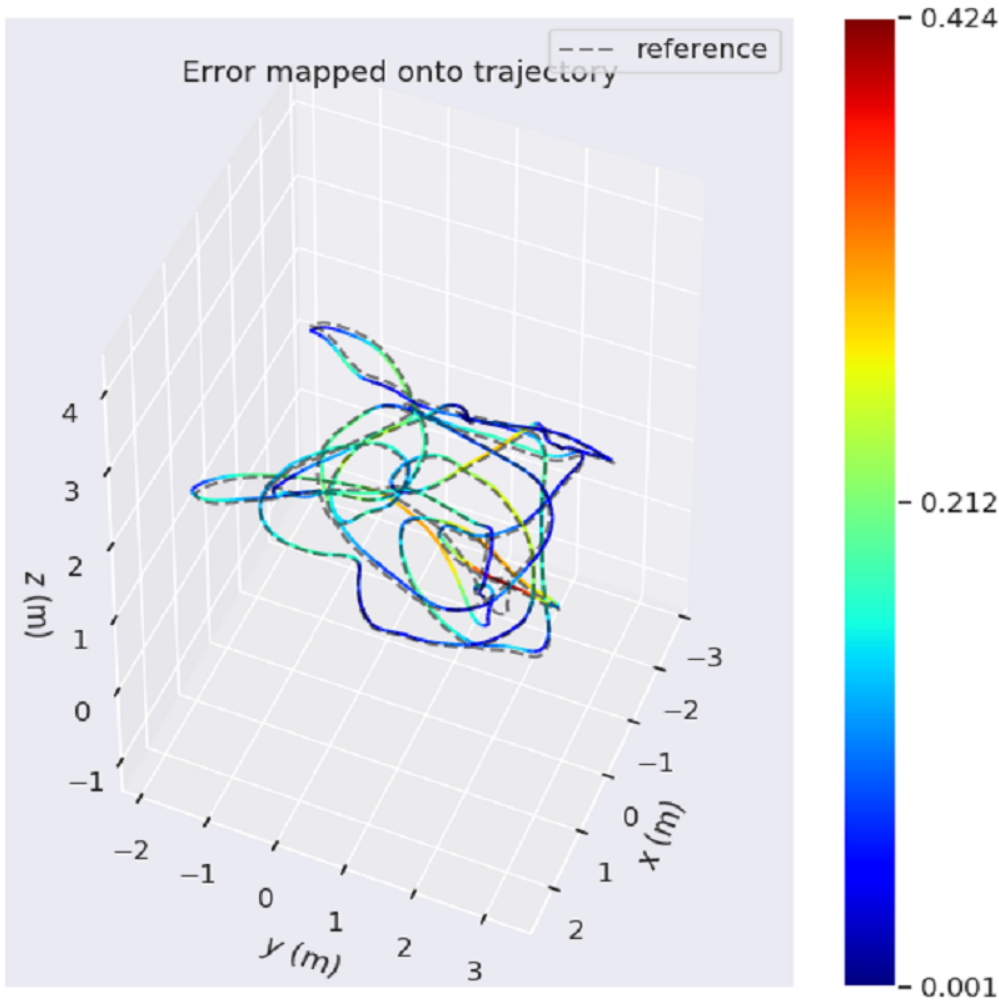}}
	\subfloat[]{\includegraphics[width=0.25\linewidth]{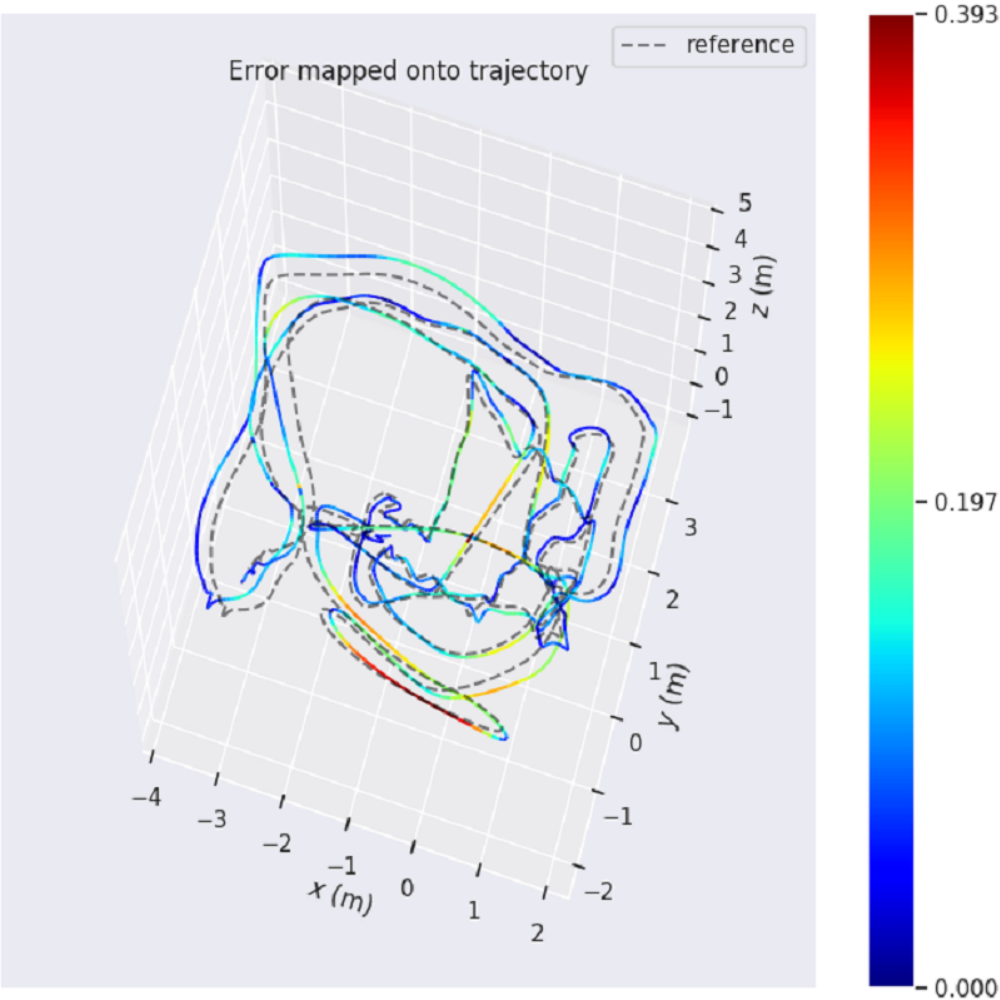}}
	\vspace{-0.1in}
	\subfloat[]{\rotatebox{90}{~~~~~~~~~~~~UV-SLAM}\includegraphics[width=0.25\linewidth]{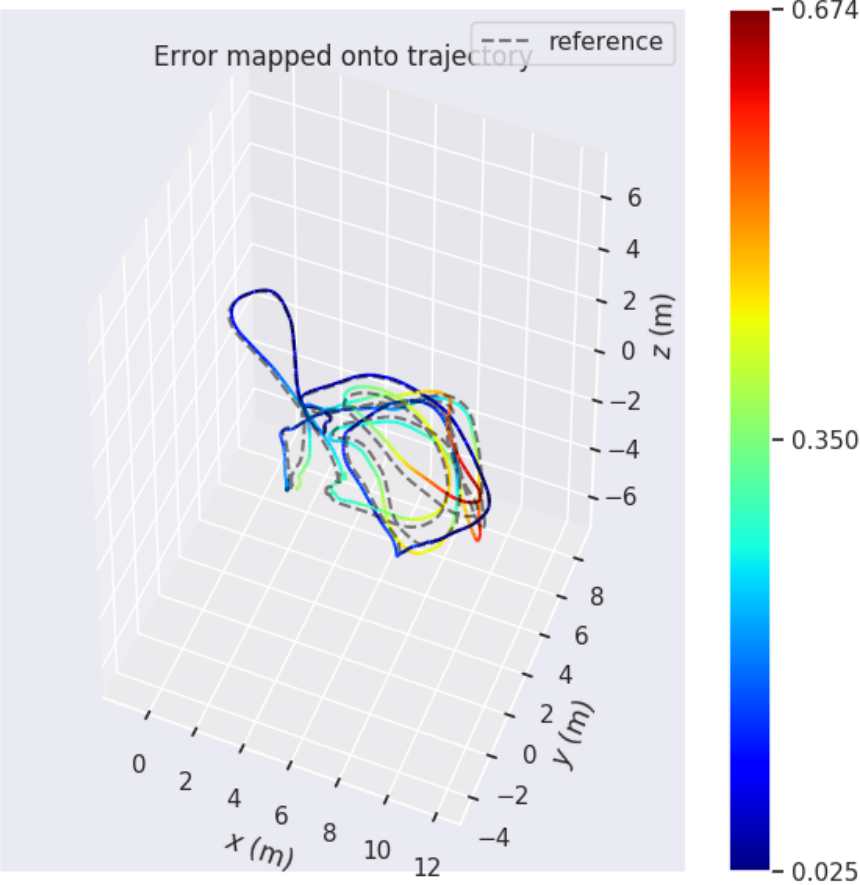}}	
	\subfloat[]{\includegraphics[width=0.25\linewidth]{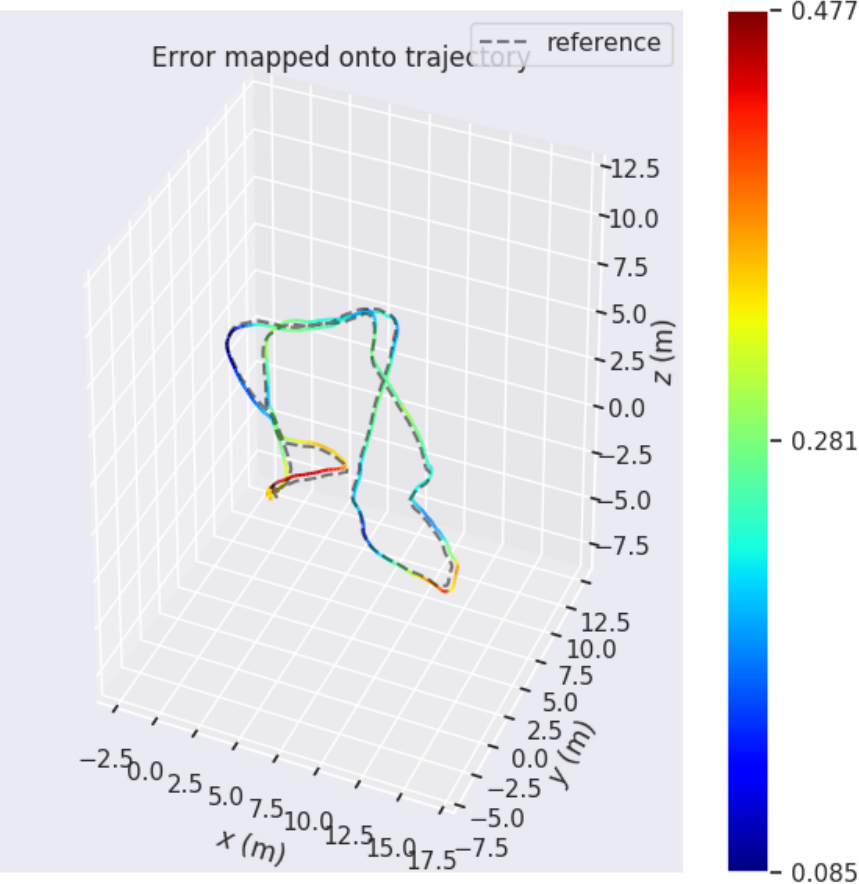}}	
	\subfloat[]{\includegraphics[width=0.25\linewidth]{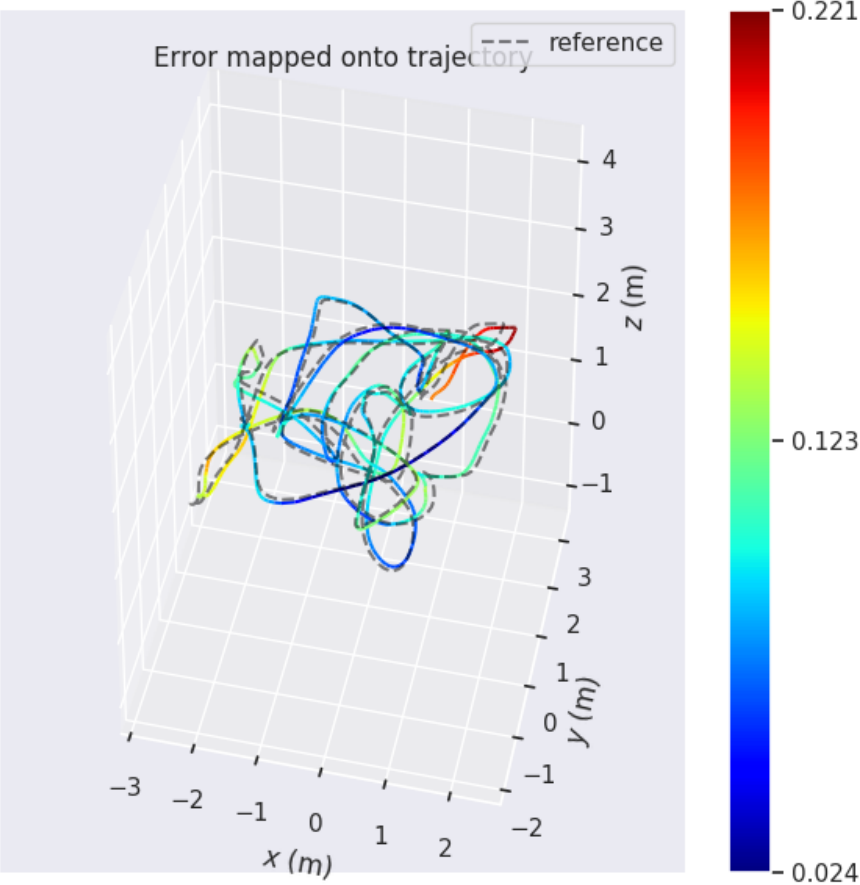}}	
	\subfloat[]{\includegraphics[width=0.25\linewidth]{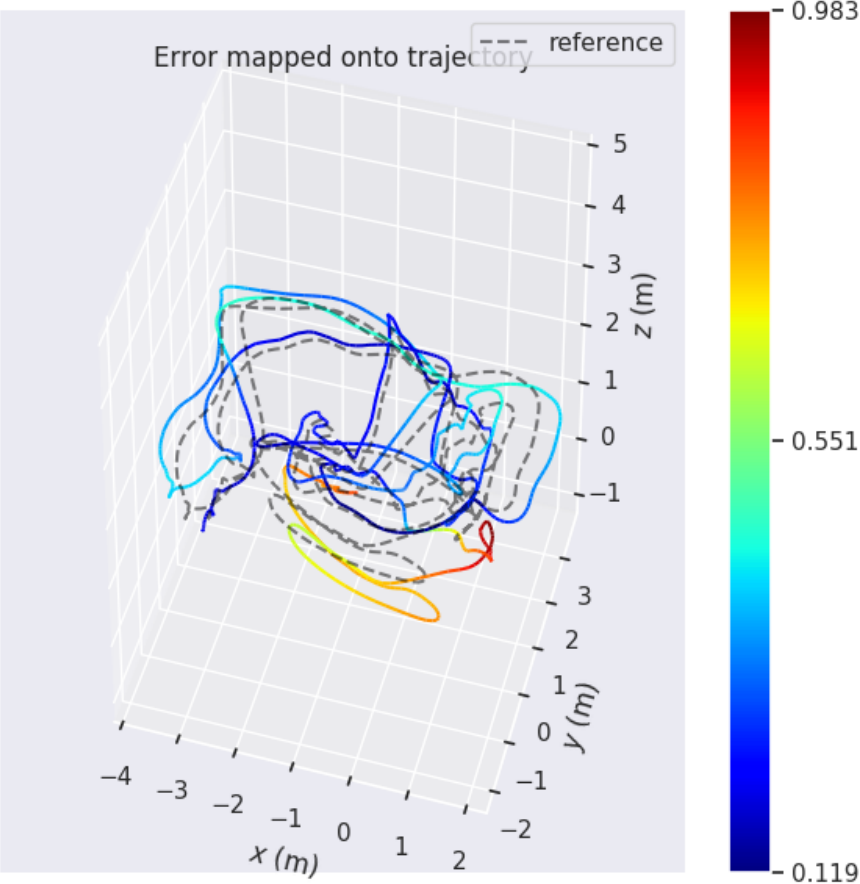}}
	\vspace{-0.1in}
	\subfloat[]{\rotatebox{90}{~~~~~~~~~~~~~~~IDLS}\includegraphics[width=0.25\linewidth]{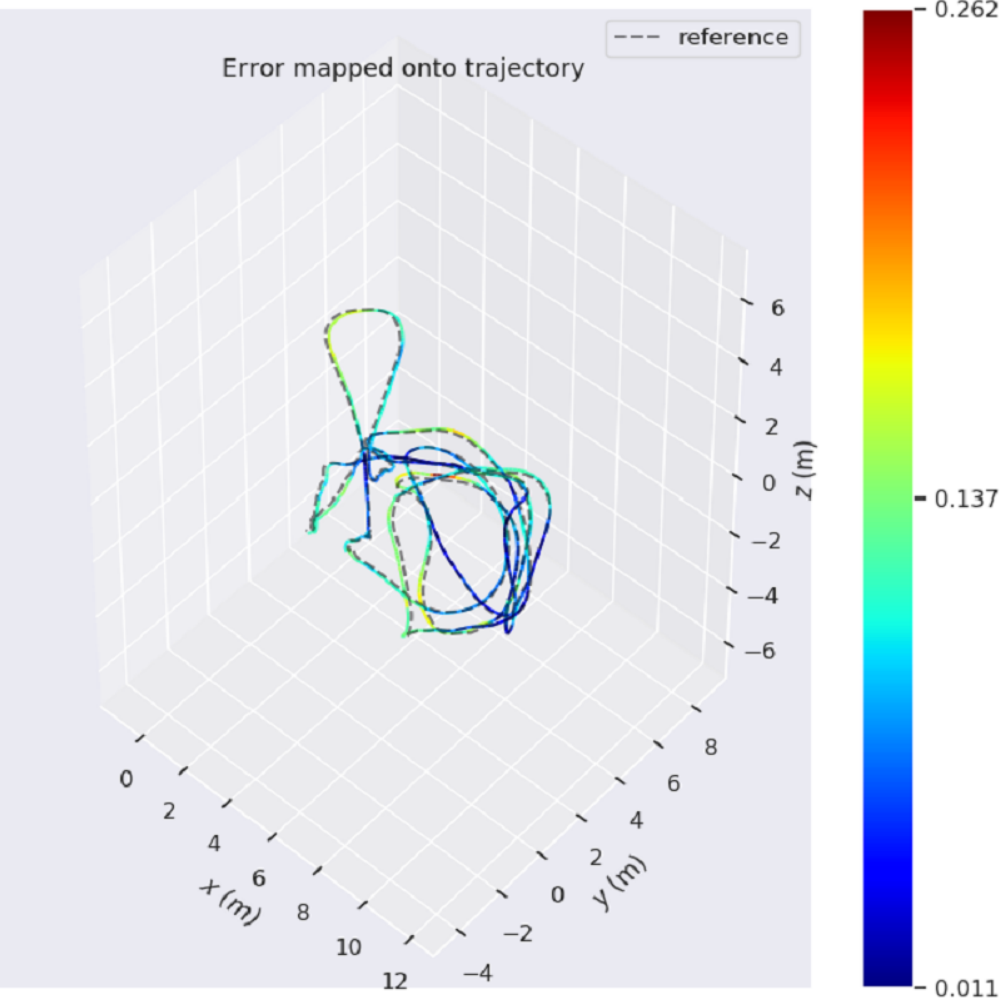}}	
	\subfloat[]{\includegraphics[width=0.25\linewidth]{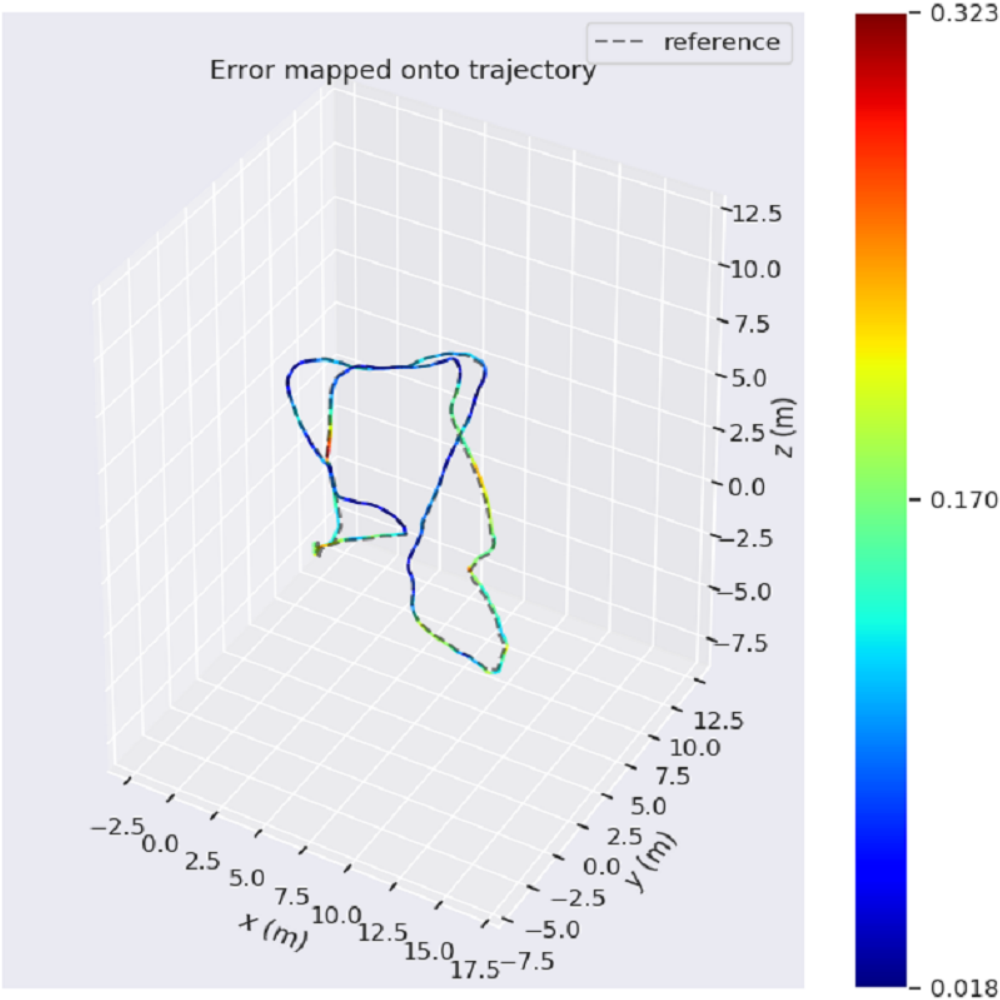}}	
	\subfloat[]{\includegraphics[width=0.25\linewidth]{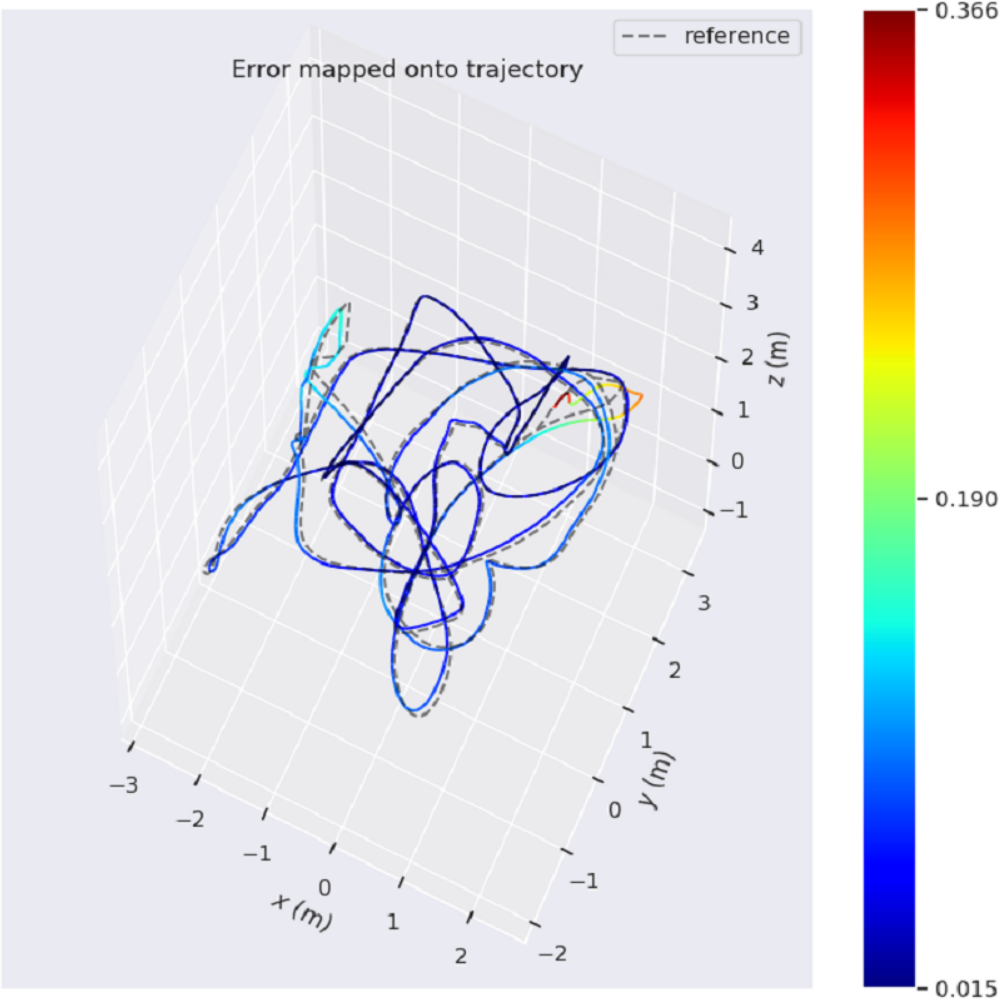}}	
	\subfloat[]{\includegraphics[width=0.25\linewidth]{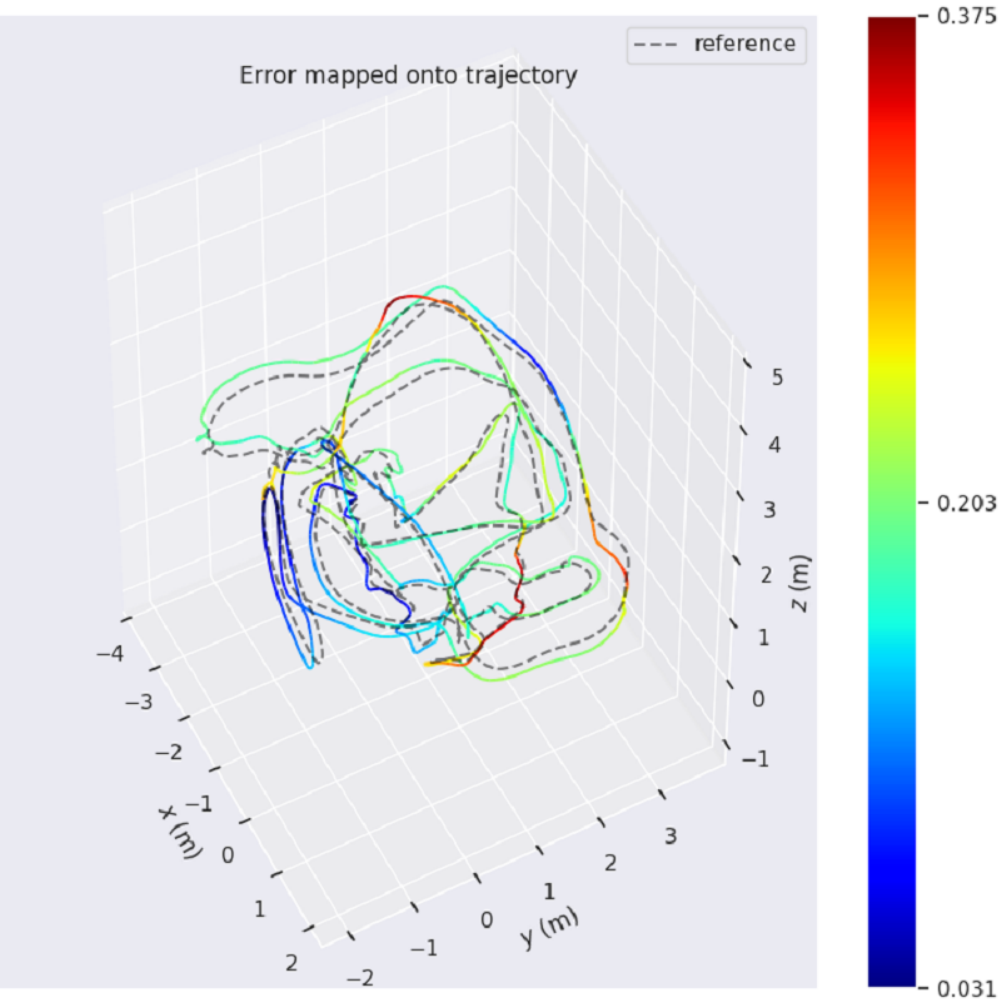}}
	\vspace{-0.1in}
	\caption{Comparison of the proposed method and other VIO systems. VINS-Mono is used to run the first row's four colorful trajectories on the (a) MH-03-medium; (b) MH-05-difficult; (c) V1-02-medium; and (d) V2-03-difficult sequences. The second row is run with PL-VIO and the third row is PL-VINS. The fourth row is run with UV-SLAM and the last row is our method(IDLS). Colors encodethe corresponding absolute pose errors.}	
	\label{fig_trajectory}
\end{figure*}

	\textbf{Visualization}: 
	To demonstrate the results intuitively, Fig.\ref{fig_trajectory} shows several trajectories estimated by IDLS and other methods. The trajectories are colored by heat-map of translation errors. The rapid camera rotations and challenging lighting conditions in these sequences make it difficult to accurately track the point features. The redder the color in the results, the higher the translation error. Comparing trajectories obtained by other four methods, we can conclude that IDLS with line features gives the smallest errors comparing with that of the other methods. In MH-03-Medium, MH-05-Difficult, V1-02-Medium, the advantage of IDLS is obvious. All these five methods don't perform well in V2-03-Difficult. IDLS still provides the minimum average translation errors.

	\subsubsection{\textbf{Comparison of Computation Efficiency}}

	We compare the computation times of different methods in this subsection. We also use the EuRoc dataset to evaluate the results.  The running times of  two threads, i.e., (1) Line Detection and Tracking and (2) VIO thread, are compared for the four methods. The results are listed in Table~\ref{tab:thread}.
	
	In Thread 1 (Line Detection and Tracking), IDLS and PL-VINS each require $\sim46ms$, while PL-VIO requires $\sim 100ms$.  UV-SLAM need more time  to detect the parallel lines and to calculate the vanishing point. In Thread 2(VIO), IDLS is more efficient than PL-VINS, which has the lowest 36.4 ms. 
	
	
	
	We also compare the number of line features, the number of parameters and execution time in Table~\ref{tab:time}. IDLS can detect more lines in each frame (around 100) than that of PL-VIO and PL-VINS (around 50) and the detected line number is close with UV-SLAM. But IDLS optimizes using fewer parameters, so the average execution time is much less than UV-SLAM. So in overall IDLS is not only more accurate, but also is more efficient.
	
	\begin{table*}[]
		\centering
		\small
		\caption{Thread Time(ms) Comparison of different methods}
		\resizebox{\textwidth}{!}{	
			\begin{tabular}{|c|cc|cc|cc|cc|}
				\hline
				\multirow{2}{*}{Method} & \multicolumn{2}{c|}{PL-VIO}                          & \multicolumn{2}{c|}{PL-VINS}                         &
				\multicolumn{2}{c|}{UV-SLAM}                         & \multicolumn{2}{c|}{IDLS}                            \\ \cline{2-9} 
				& \multicolumn{1}{c|}{Thread 1 Time} & Thread 2  Time& \multicolumn{1}{c|}{Thread 1 Time} & Thread 2  Time &
				\multicolumn{1}{c|}{Thread 1 Time} & Thread 2  Time& \multicolumn{1}{c|}{Thread 1 Time} & Thread 2  Time \\ \hline
				Mean (ms)                   & \multicolumn{1}{c|}{100}            & 48             & \multicolumn{1}{c|}{46}             & 46             &\multicolumn{1}{c|}{63}             & 69             & \multicolumn{1}{c|}{\textbf{46}}             & \textbf{36.4}           \\ \hline
			\end{tabular}
		}
		\label{tab:thread}
	\end{table*}
	
	\begin{table*}[]
		\centering
		\normalsize
		\caption{Average Execution Time(ms) Comparison of different methods}
		\resizebox{\textwidth}{!}{
			\begin{tabular}{|c|ccc|ccc|ccc|ccc|}
				\hline
				\multirow{2}{*}{Method} & \multicolumn{3}{c|}{PL-VIO}                                              & \multicolumn{3}{c|}{PL-VINS}                                             &
				\multicolumn{3}{c|}{UV-SLAM} & \multicolumn{3}{c|}{IDLS}                                                \\ \cline{2-13} 
				& \multicolumn{1}{c|}{Line Num} & \multicolumn{1}{c|}{Line Params} & Times & \multicolumn{1}{c|}{Line Num} & \multicolumn{1}{c|}{Line Params} & Times & \multicolumn{1}{c|}{Line Num} & \multicolumn{1}{c|}{Line Params} & Times &\multicolumn{1}{c|}{Line Num} & \multicolumn{1}{c|}{Line Params} & Times \\ \hline
				Mean                    & \multicolumn{1}{c|}{$\sim 5\times 10^1$}       & \multicolumn{1}{c|}{4}           & 48    & \multicolumn{1}{c|}{$\sim 5\times 10^1$}       & \multicolumn{1}{c|}{4}           & 46    &\multicolumn{1}{c|}{$\sim 1\times 10^2$}       & \multicolumn{1}{c|}{7}           & 69& \multicolumn{1}{c|}{\textbf{$\sim 1\times 10^2$}}      & \multicolumn{1}{c|}{\textbf{2}}           & \textbf{36.4}\color{red}$( \downarrow 20.87\% )$  \\ \hline
			\end{tabular}
		}
		\label{tab:time}
	\end{table*}

	\subsection{Real Environment Dataset}
	We in specially collect a dataset in challenging indoor environments to test different algorithms' ability to handle the weak-textured environments and long straight corridors. It comprises of  12 trajectories. The environments of collection of these trajectories are introduced in the second column of Table~\ref{tab:real}. 
	By evaluating the performances of different algorithms on these challenging trajectories, we focus on comparing the robustness and accuracy of the different methods.
	
	\begin{table*}[]
		\centering
		\small
		\caption{Mean ATE [M] Comparison of different methods}
		\begin{tabular}{|c|c|c|c|c|c|c|}
			\hline
			Sequence & Environment                                                                                & Vins-Mono & PL-VIO & PL-VINS & UV-SLAM & IDLS  \\ \hline
			01       & \multirow{3}{*}{Lab}                                                                       & 0.060     & 0.054  & 0.052   & 0.048   & \textbf{0.043} \\ \cline{1-1} \cline{3-7} 
			02       &                                                                                            & 0.073     & 0.063  & 0.059   & 0.051   & \textbf{0.045} \\ \cline{1-1} \cline{3-7} 
			03       &                                                                                            & 0.046     & 0.038  & 0.032   & 0.023   & \textbf{0.022} \\ \hline
			04       & Pantry(dim light)                                                                          & failed    & 0.232  & 0.192   & 0.191   & \textbf{0.183} \\ \hline
			05       & \multirow{2}{*}{\begin{tabular}[c]{@{}c@{}}Corridor  (blank/glass wall)\end{tabular}}     & failed    & failed & failed  & 0.268   & \textbf{0.264} \\ \cline{1-1} \cline{3-7} 
			06       &                                                                                            & failed    & failed & failed  & 0.184   & \textbf{0.139} \\ \hline
			07       & \multirow{2}{*}{\begin{tabular}[c]{@{}c@{}}Lab+Corridor  (blank/glass wall)\end{tabular}} & failed    & failed & 1.425   & 0.232   & \textbf{0.163} \\ \cline{1-1} \cline{3-7} 
			08       &                                                                                            & failed    & 1.735  & 0.523   & 0.265   & \textbf{0.257} \\ \hline
			09       & \begin{tabular}[c]{@{}c@{}}Mail room  (similar feature)\end{tabular}                      & 3.523     & 0.824  & 0.856   & 1.134   & \textbf{0.128} \\ \hline
			10       & \multirow{2}{*}{Mail room+Hall}                                                            & 0.440     & 0.245  & 0.241   & 0.263   & \textbf{0.143} \\ \cline{1-1} \cline{3-7} 
			11       &                                                                                            & 0.317     & 0.213  & 0.182   & 0.155   & \textbf{0.134} \\ \hline
			12       & \begin{tabular}[c]{@{}c@{}}Corridor(doors)  +Hall\end{tabular}                            & 0.287     & 0.183  & 0.173   & 0.136   & \textbf{0.124} \\ \hline
		\end{tabular}
		\label{tab:real}
	\end{table*}
	
\begin{figure}[!htbp]
		\centering
		\includegraphics[width=1\linewidth]{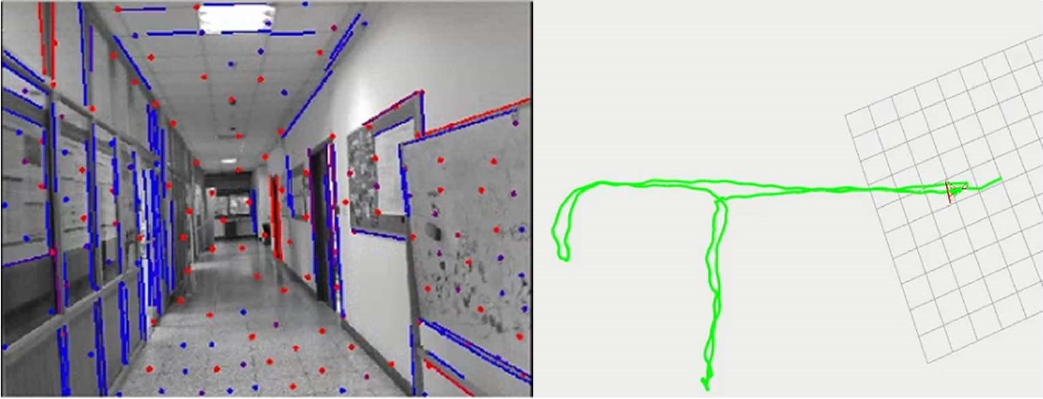}
		\caption{Visualization result of sequence-06 for IDLS. }
		\vspace{-0.3in}
		\label{fig:visual}
	\end{figure}

	The evaluation results are shown in the Table~\ref{tab:real}. It can be seen that PL-VIO and PL-VINS are not able to initialize in the 05 and 06 sequences when following a very straight corridor (i.e. a degraded line feature environment). The number of line features is insufficient and the trajectory tracking fails. UV-SLAM, on the other hand, uses vanishing point matching as an additional condition, can solve this problem relatively well. The IDLS uses a two-point inverse depth line representation, an improved triangulation method and a reprojection error model. It allows the trajectory to be optimised without introducing erroneous line direction line features. It can be seen that IDLS  runs successfully and is more robustly in all the challenging sequences.
	As summarized in Table~\ref{tab:real}, the errors of the IDLS in all the 12 trajectories are lower than the other SOTA methods. The visual result is shown in Fig~\ref{fig:visual}.
	
	\subsection{Discussion}
	As can be seen from the experiments, the inverse-depth line representation  effectively improves the accuracy and efficiency of the  optimization process. At the same time, the proposed multi-frame optimization-based line triangulation method allows more lines to be initialized. The proposed IDLS is more accurate and efficient than the SOTA method UV-SLAM.

    However, from Proposition 2, IDLS cannot initialize lines in the case of complete degradation. But such cases appear rarely in experiments. 
    The reduced dimension representation hasn't stored the lines, which doesn't affect the camera localization but affects the reconstruction of 3D maps. 
    We can introduce addition operations to store lines and add more restrictions on the structure of the lines in the next work in order to improve the reconstruction of feature maps.
	\color{black}

	\section{Conclusion}
	
	In this paper, 
	we propose to represent each line feature by the inverse depths of two end points of the line. This compact representation  reduces the parameter numbers during initialization and optimization. 
	We in particular present the inverse depth based initialization, residue function setup, and sliding window based optimization process. 
	The Jacobian matrix of the line feature residual errors is given in detail in order to efficiently solve the optimization problem. 
	A two-step optimization method is proposed to solve the optimization problem in the sliding window. 
	We compared IDLS method with four state-of-the-art PL-SLAM methods. 
	The results show that the inverse depth line representation can not only help to  improve the  accuracy in challenging environments but also can improve the efficiency to meet the real-time requirements. 
	In the future, we will investigate line feature-based loop closure and indoor map reconstruction by introducing the inverse depth line features in 3D space.

	\bibliographystyle{IEEEtran}
	\bibliography{references.bib}

		\newpage
		\begin{IEEEbiography}
			[{\includegraphics[width=1in,height=1in,clip,keepaspectratio]{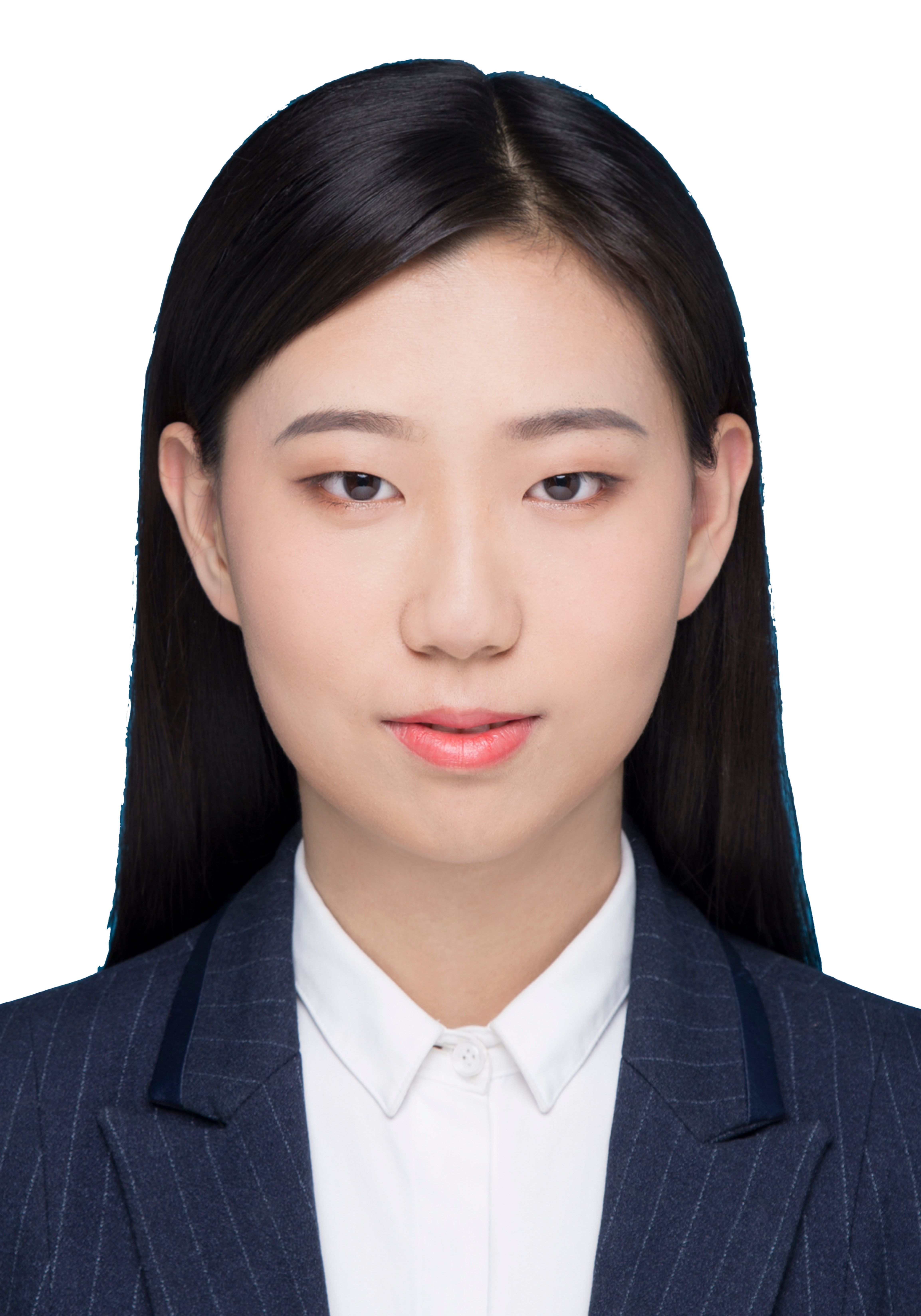}}]{Wanting Li}is a PhD student in the Department of Computer Science, Renmin University of China. She received the BS degree in Computer science and technology from China Agricultural University (2019). Her research is indoor locating algorithms and unsupervised learning in pervasive computing. 
		\end{IEEEbiography}
		\vskip -2\baselineskip plus -1fil
		\begin{IEEEbiography}
			[{\includegraphics[width=1in,height=1in,clip,keepaspectratio]{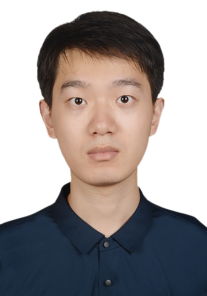}}]{Shuo Wang}  received the BS degree from the School of Information, Renmin University of China, where he is pursuing the PhD degree. His research interests include network localization algorithms, graph optimization, and applications.
		\end{IEEEbiography}
		\vskip -2\baselineskip plus -1fil
		\begin{IEEEbiography}
			[{\includegraphics[width=1in,height=1in,clip,keepaspectratio]{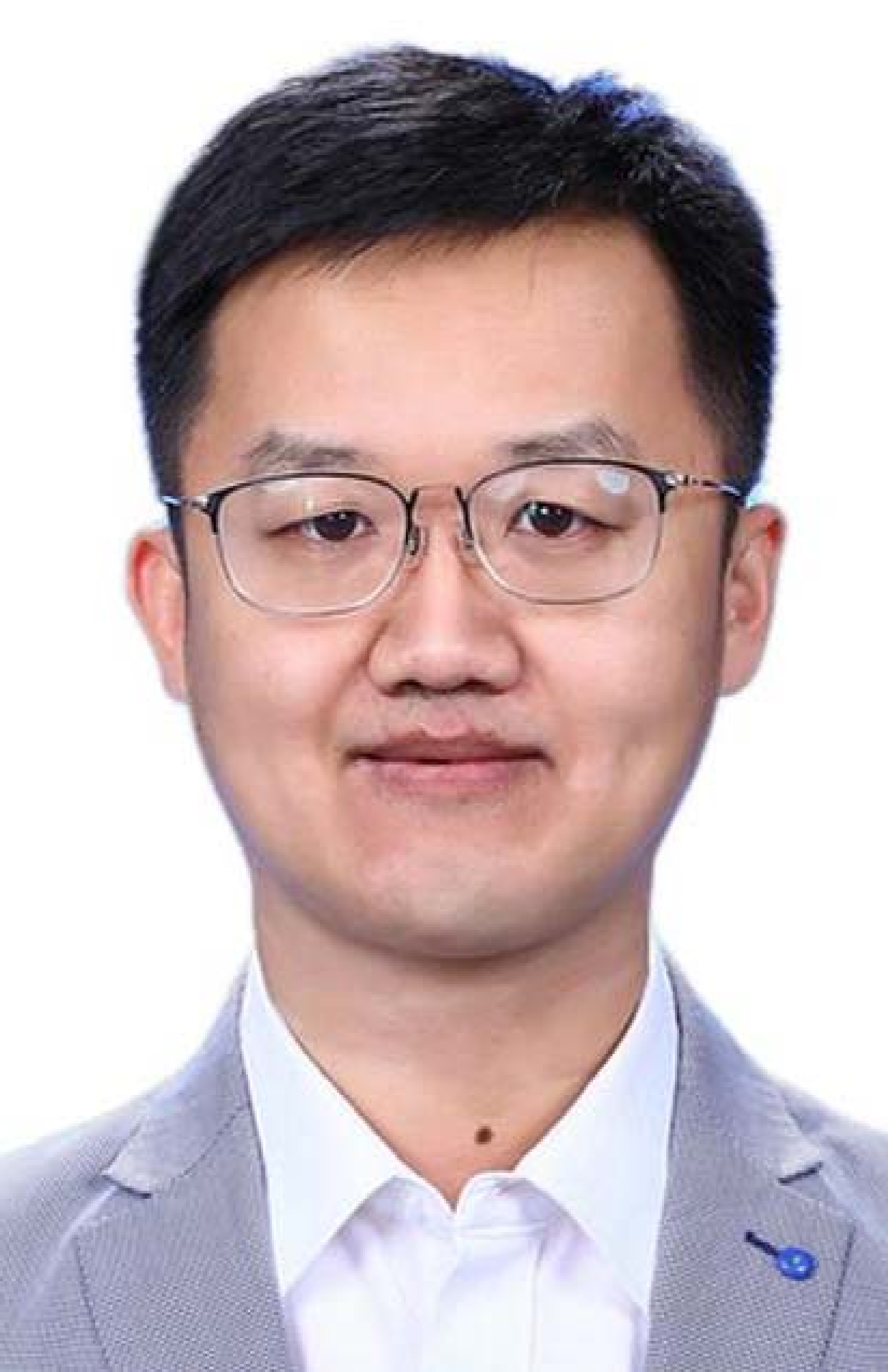}}]{Yongcai Wang}
			received his BS and PhD degrees from department of automation sciences and engineering, Tsinghua University in 2001 and 2006 respectively. He worked as associated researcher at NEC Labs. China from 2007-2009. He was an research scientist in Institute for Interdisciplinary Information Sciences (IIIS), Tsinghua University from 2009-2015. He was a visiting scholar at Cornell University in 2015. He is currently associate professor at  Department of Computer Sciences, Renmin University of China.  His research interests include network localization and crowdsourcing based indoor location and mapping.   
		\end{IEEEbiography}
       	\vskip -2\baselineskip plus -1fil
       \begin{IEEEbiography}
       	[{\includegraphics[width=1in,height=1in,clip,keepaspectratio]{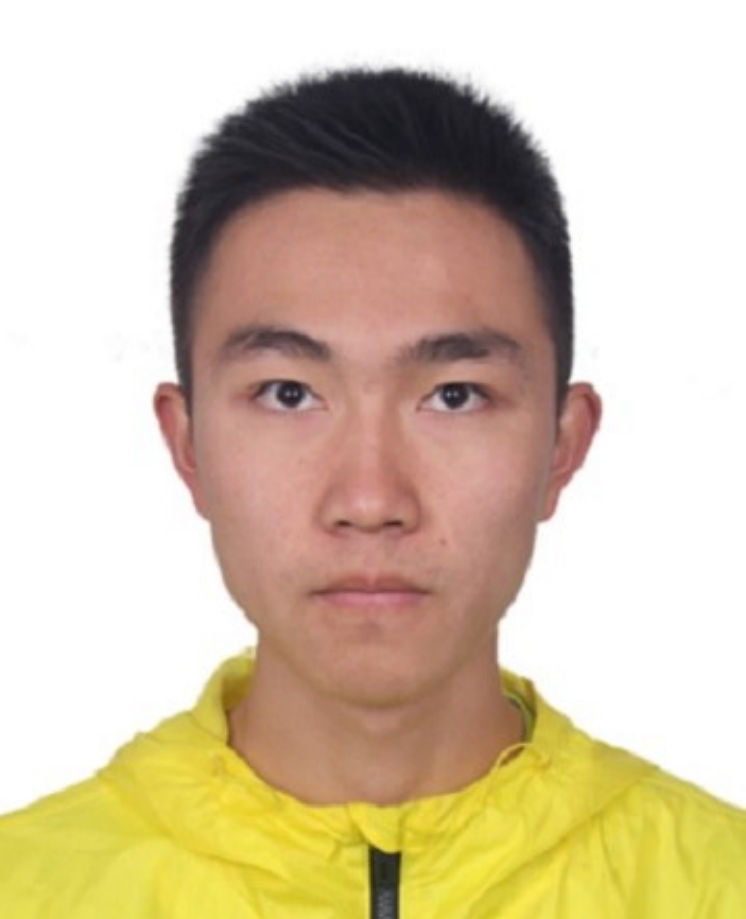}}]{Yu Shao} is a master student in the Department of Computer Science, Renmin University of China.
       	He received the BS degree in Computer science and technology from Renmin University of China (2021). His research interest is lidar SLAM (Simultaneous Localization and Mapping) algorithms.
       \end{IEEEbiography}
		\vskip -2\baselineskip plus -1fil
		\begin{IEEEbiography}[{\includegraphics[width=1in,height=1.1in,clip,keepaspectratio]{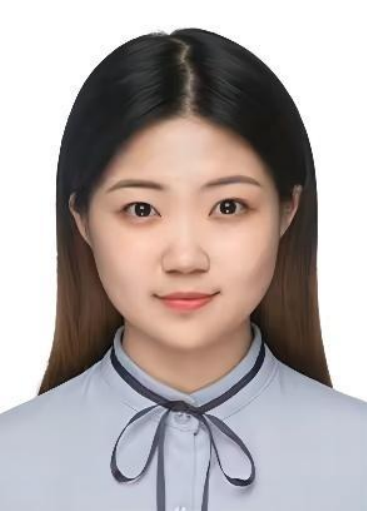}}]{Xuewei Bai} received the BS degree
			from the Department of Computer Science and
			Technology, North China Electric Power University, in 2020. She is currently working toward the PhD degree in the Department of Computer Sciences, Renmin University of China. Her research interests include network localization algorithms and graph optimization and applications.	
		\end{IEEEbiography}
		\vskip -2\baselineskip plus -1fil
		\begin{IEEEbiography}
			[{\includegraphics[width=1in,height=1in,clip,keepaspectratio]{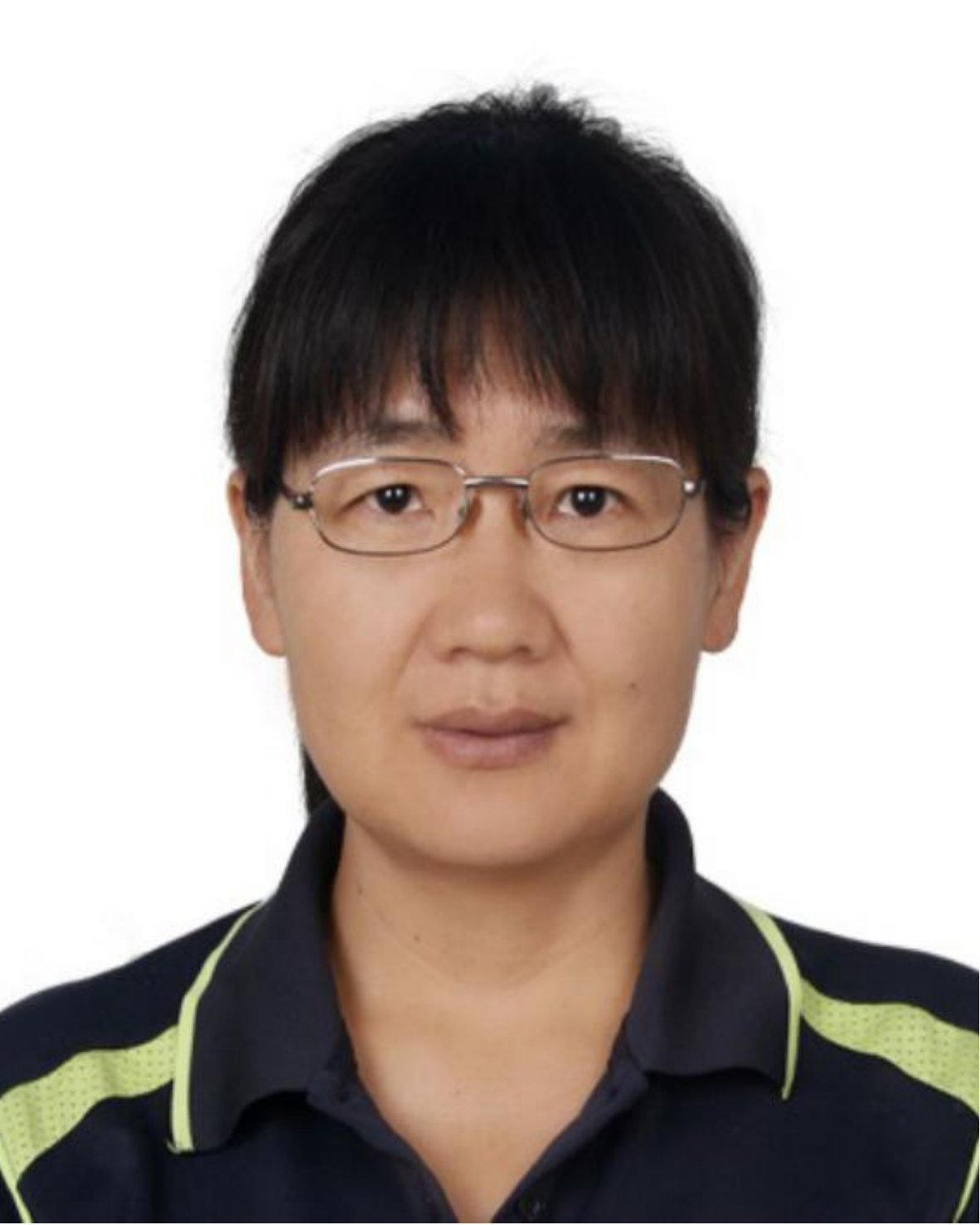}}]{Deying Li}
			received the MS degree in Mathematics from Huazhong Normal University (1988)
			and PhD degree in Computer Science from City University of Hong Kong (2004). She is 
			currently a Professor in the Department of Computer Science, Renmin University of China. Her
			research includes wireless networks, mobile computing, social network and algorithm design and analysis. 
		\end{IEEEbiography}
		
	\end{document}